\documentclass{article}
\usepackage[preprint]{colm2026_conference}

\usepackage{microtype}
\usepackage{hyperref}
\usepackage{url}
\usepackage{booktabs}
\usepackage{graphicx}
\usepackage{multirow}
\usepackage{xcolor}
\usepackage{amsmath}

\newcommand{\ci}[2]{\,{\scriptsize[#1,\,#2]}}

\definecolor{darkblue}{rgb}{0, 0, 0.5}
\hypersetup{colorlinks=true, citecolor=darkblue, linkcolor=darkblue, urlcolor=darkblue}

\title{Detecting and Correcting Reference Hallucinations in Commercial LLMs and Deep Research Agents}

\author{Delip Rao\thanks{Corresponding author} \\
  University of Pennsylvania \\
  \texttt{delip@seas.upenn.edu} \\
  \And
  Eric Wong \\
  University of Pennsylvania \\
  \texttt{exwong@seas.upenn.edu} \\
  \And
  Chris Callison-Burch \\
  University of Pennsylvania \\
  \texttt{ccb@seas.upenn.edu}
}

\begin{document}

\maketitle

\begin{abstract}
  Large language models and deep research agents supply citation URLs to support their claims, yet the reliability of these citations has not been systematically measured. We address six research questions about citation URL validity using 10 models and agents on DRBench (53,090 URLs) and 3 models on ExpertQA (168,021 URLs across 32 academic fields). We find that 3--13\% of citation URLs are hallucinated---they have no record in the Wayback Machine and likely never existed---while 5--18\% are non-resolving overall. Deep research agents generate substantially more citations per query than search-augmented LLMs but hallucinate URLs at higher rates. Domain effects are pronounced: non-resolving rates range from 5.4\% (Business) to 11.4\% (Theology), with per-model effects even larger. Decomposing failures reveals that some models fabricate every non-resolving URL, while others show substantial link-rot fractions indicating genuine retrieval. As a solution, we release \texttt{urlhealth}, an open-source tool for URL liveness checking and stale-vs-hallucinated classification using the Wayback Machine. In agentic self-correction experiments, models equipped with \texttt{urlhealth} reduce non-resolving citation URLs by $6\textrm{--}79\times$ to under 1\%, though effectiveness depends on the model's tool-use competence. The tool and all data are publicly available. Our characterization findings, failure taxonomy, and open-source tooling establish that citation URL validity is both measurable at scale and correctable in practice.
\end{abstract}

%% ============================================================
%% SECTION 1: INTRODUCTION
%% ============================================================
\section{Introduction}
\label{sec:introduction}

Retrieval-augmented generation (RAG) and web search integration are now standard across all major LLM providers, and every provider offers a ``deep research'' mode that generates long-form reports with inline citations. Users treat these citations as evidence. However, citation hallucination by large language models has already caused real-world harm in several notable incidents ranging from legal sanctions to academic retractions to medical misinformation\footnote{In \href{https://www.nytimes.com/2023/06/08/nyregion/lawyer-chatgpt-sanctions.html}{\textit{Mata v. Avianca} (2023)}, a New York attorney was sanctioned after submitting a brief that contained fabricated case citations generated by \texttt{ChatGPT}. A paper co-authored by Tsinghua University researchers was \href{https://www.theregister.com/2024/02/01/iclr_paper_ai_references/}{withdrawn from ICLR} after reviewers identified AI-generated references. \href{https://unchartedterritories.tomaspueyo.com/p/perplexity-is-lying}{Perplexity AI fabricated reviews} attributed to named medical doctors.}. Yet no systematic study has characterized the reliability of LLM-generated citation URLs across models, providers, and knowledge domains, and no reusable tools exist for detecting and mitigating broken and hallucinated URLs.

We organize this study around six research questions:

\begin{description}
  \item[\textbf{RQ1} (Prevalence)] How prevalent are citation URL failures across commercial LLMs and deep research agents?
  \item[\textbf{RQ2} (System type)] How do deep research agents compare with search-augmented LLMs?
  \item[\textbf{RQ3} (Domain variation)] How strongly does citation reliability vary across academic fields?
  \item[\textbf{RQ4} (Failure composition)] What fraction of citation failures is fabrication versus link rot?
  \item[\textbf{RQ5} (Quantity vs.\ quality)] Does producing more citations improve per-citation reliability?
  \item[\textbf{RQ6} (Mitigation)] Can post-hoc URL verification reduce citation hallucination?
\end{description}

Our main findings are: 1) \textbf{RQ1:} Hallucinated URL rates range from 3\% to 13\%, with 5--18\% non-resolving overall. 2) \textbf{RQ2:} Deep research agents exhibit the highest hallucination rates despite generating far more citations; retrieval architecture matters more than output volume. 3) \textbf{RQ3:} Non-resolving rates vary from 5.4\% (Business) to 11.4\% (Theology), with per-model effects reaching 4.3$\times$. 4) \textbf{RQ4:} Some models fabricate every non-resolving URL, while others show substantial stale fractions indicating genuine retrieval. 5) \textbf{RQ5:} More citations per query do not mean fewer errors per citation. 6) \textbf{RQ6:} In agentic self-correction experiments, \texttt{urlhealth} reduces non-resolving URLs by $6\textrm{--}79\times$ (all $p < 10^{-35}$), with post-mitigation rates below 1\% across all three models, though effectiveness depends on tool-use competence.

\section{Definitions and Scope}
\label{sec:definitions}

A \textbf{citation} in the context of LLM-generated text refers to a URL, bibliographic reference, or attributed snippet intended to support a claim. This work focuses on URL-based citations, which we classify as follows. A \textbf{non-resolving URL} is one that returns an HTTP error (4xx or 5xx status code) or fails to connect. This is the broadest class of unreliable citations. A \textbf{hallucinated URL} is a non-resolving URL for which no archived snapshot exists in the Wayback Machine~\footnote{https://web.archive.org/} at any point in time. That is, there is no historical evidence the URL ever existed.\footnote{This is an operational definition. Because Wayback coverage is substantial but incomplete and non-uniform \citep{ainsworth_how_2013,alsum_profiling_2013,alkwai_how_2015}, our estimates of hallucinated URL rates should be interpreted as conservative lower bounds. Three factors contribute: (1) URLs that never existed cannot appear in the Wayback Machine, but URLs that did exist may also be absent if they were never crawled, so absent Wayback snapshots under-count stale URLs and over-count hallucinated ones; (2) our pipeline excludes HTTP~403 responses (bot-blocking), which a headless-browser audit confirms are 99.7\% live (Appendix~\ref{app:unknown-audit}); (3) the residual UNKNOWN category (non-200, non-404 responses) is 89\% live or blocked when probed with a real browser. The only plausible over-counting direction---Wayback archiving soft-404 or wildcard-redirect pages---is small relative to these under-counting factors.} Hallucinated URLs form a strict subset of non-resolving URLs. A \textbf{stale URL} is a non-resolving URL that does have a Wayback Machine snapshot---it existed at some point but is now unreachable. Stale URLs represent natural link rot rather than fabrication.

The relationship among these categories can be formalized as
\[
\text{Hallucinated URLs} = \text{Non-resolving URLs} - \text{Stale URLs}
\]

\paragraph{Scope.} This study focuses on URL-based citation hallucinations. Two additional failure modes---fabricated snippets (quoted text not appearing at the cited page) and invented bibliographic entries (plausible but incorrect metadata such as wrong authors or non-existent DOIs)---are documented qualitatively but require separate systematic study.

%% ============================================================
%% SECTION 3: EXPERIMENTAL SETUP
%% ============================================================
\section{Experimental setup}
\label{sec:setup}

\subsection{Datasets}
\label{sec:datasets}

We use two complementary datasets. \textbf{DRBench} \citep{du2025drbench} comprises 100 multilingual research queries (Chinese and English) covering finance, science, and technology, with pre-collected outputs from 23 models. \textbf{ExpertQA} \citep{malaviya2024expertqa} consists of 2,177 expert-curated questions across 32 academic and professional fields. DRBench enables cross-model comparison of citation URL validity; ExpertQA enables domain-stratified analysis of fabrication rates. A dataset summary table is provided in Appendix~\ref{app:experimental}.

\subsection{Models}
\label{sec:models}

Table~\ref{tab:models} summarizes the models we evaluate. From the 23 models in DRBench, we analyze 10 from three providers (Google, OpenAI, Anthropic) for which reliable URL liveness data is available; exclusion criteria for the remaining 13 are detailed in Appendix~\ref{app:experimental}.

We distinguish two model types. \textit{Deep research agents} (\texttt{gemini-2.5-pro-deepresearch}, \texttt{openai-deepresearch}) execute multi-step retrieval and synthesis, producing long-form reports. \textit{Search-augmented LLMs} perform a single query with search integration. For ExpertQA, we evaluate three search-augmented models---\texttt{claude-sonnet-4-5}, \texttt{gemini-2.5-pro}, and \texttt{gpt-5.1}---each prompted to return structured JSON with a markdown answer and numbered citations. Provider-specific API configurations are listed in Appendix~\ref{app:experimental}.

\begin{table}[t]
\centering
\small
\begin{tabular}{llllr}
\toprule
\textbf{Model} & \textbf{Provider} & \textbf{Type} & \textbf{Study} & \textbf{URLs} \\
\midrule
\texttt{gemini-2.5-pro-deepresearch} & Google & Deep research & DRBench & 11,309 \\
\texttt{openai-deepresearch} & OpenAI & Deep research & DRBench & 4,121 \\
\midrule
\texttt{gemini-2.5-pro-with-search} & Google & Search-aug. & DRBench & 1,609 \\
\texttt{gemini-2.5-flash-with-search} & Google & Search-aug. & DRBench & 2,433 \\
\texttt{gpt-4o-search-preview} & OpenAI & Search-aug. & DRBench & 387 \\
\texttt{gpt-4o-mini-search-preview} & OpenAI & Search-aug. & DRBench & 402 \\
\texttt{gpt-4.1} & OpenAI & Search-aug. & DRBench & 336 \\
\texttt{gpt-4.1-mini} & OpenAI & Search-aug. & DRBench & 296 \\
\texttt{claude-3-7-sonnet-with-search} & Anthropic & Search-aug. & DRBench & 1,735 \\
\texttt{claude-3-5-sonnet-with-search} & Anthropic & Search-aug. & DRBench & 641 \\
\midrule
\texttt{claude-sonnet-4-5} & Anthropic & Search-aug. & ExpertQA & 61,407 \\
\texttt{gemini-2.5-pro} & Google & Search-aug. & ExpertQA & 22,878 \\
\texttt{gpt-5.1} & OpenAI & Search-aug. & ExpertQA & 83,736 \\
\bottomrule
\end{tabular}
\caption{Model inventory. ``Search-aug.'' denotes search-augmented LLMs. URL counts reflect total unique URLs extracted from model outputs.}
\label{tab:models}
\end{table}

\subsection{URL extraction and classification}
\label{sec:url-extraction}
\label{sec:url-collection}
\label{sec:dead-url-detection}
\label{sec:hallucinated-url-detection}

For DRBench, URLs are extracted from model-generated text via regex matching of \texttt{https?://} patterns. For ExpertQA, URLs are parsed from provider-specific API response structures (Appendix~\ref{app:experimental}).

Each URL is tested with an HTTP HEAD request (falling back to GET on 405, 403, or 501 responses). URLs returning 4xx or 5xx status codes, connection errors, or timeouts are classified as \textit{non-resolving}, except 403 responses, which often reflect bot-blocking rather than genuinely dead pages.\footnote{In the ExpertQA experiment, 403 responses account for 6.6--17.0\% of all URLs depending on the model, predominantly from academic publisher domains (\texttt{sciencedirect.com}, \texttt{researchgate.net}, \texttt{mdpi.com}, \texttt{academic.oup.com}, \texttt{tandfonline.com}). This exclusion is conservative: some 403 responses may reflect truly non-resolving URLs, so our non-resolving rates are lower bounds. See Appendix~\ref{app:sensitivity} for a sensitivity analysis.} The User-Agent header mimics a standard browser to reduce false positives.

Each non-resolving URL is then checked against the Wayback Machine API.\footnote{\url{https://archive.org/help/wayback_api.php}} If no archived snapshot exists at any timestamp, the URL is classified as \textit{hallucinated}; otherwise it is classified as \textit{stale} (a real page that has since gone offline). As noted in Section~\ref{sec:definitions}, this is a lower-bound estimate of hallucination because incomplete archive coverage can misclassify some hallucinated URLs as stale. Additional pipeline details are in Appendix~\ref{app:experimental}.

%% ============================================================
%% SECTION 4: CHARACTERIZATION RESULTS
%% ============================================================
\section{Characterization results}
\label{sec:characterization}

\subsection{RQ1: How prevalent are citation URL failures?}
\label{sec:rq1}

\begin{figure}[t]
\centering
\includegraphics[width=\linewidth]{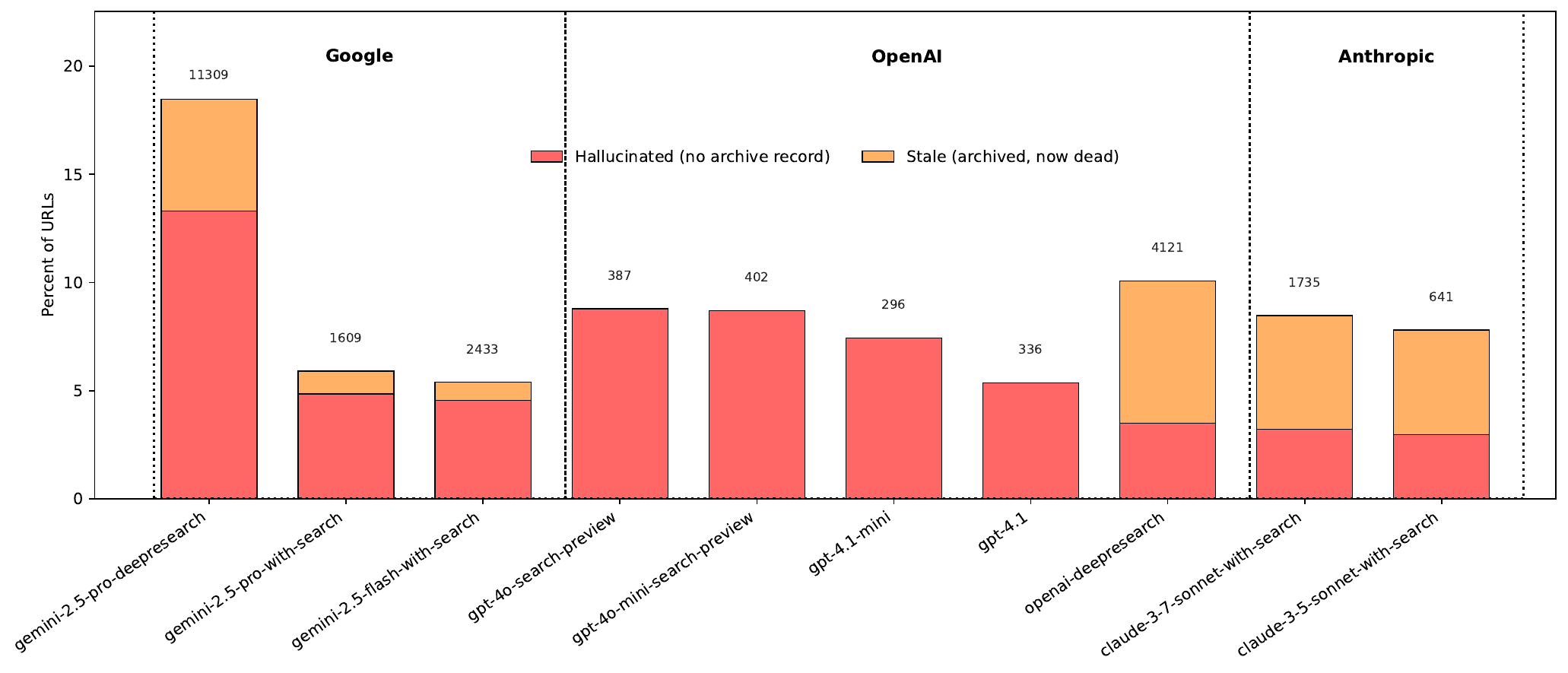}
\caption{Non-resolving URL rates for DRBench models, grouped by provider. Each bar decomposes into hallucinated URLs (red; no Wayback Machine archive, indicating the URL likely never existed) and stale URLs (orange; archived but currently dead). Numbers above bars indicate total URLs per model.}
\label{fig:drbench-barplot}
\end{figure}

Table~\ref{tab:drbench-results} and Figure~\ref{fig:drbench-barplot} report per-model URL validity results for the 10 models we analyzed on DRBench.

\paragraph{Overall rates.} Non-resolving URL rates across models range from 5.4\% [3.0, 7.7] (\texttt{gpt-4.1}) to 18.5\% [17.8, 19.2] (\texttt{gemini-2.5-pro-deepresearch}), while hallucinated URL rates range from 3.0\% [1.7, 4.4] (\texttt{claude-3-5-sonnet-with-search}) to 13.3\% [12.7, 13.9] (\texttt{gemini-2.5-pro-deepresearch}). All intervals are bootstrap 95\% CIs.

Among providers, \texttt{gemini-2.5-pro-deepresearch} has the highest hallucination rate (13.3\%) despite a top DRBench quality score; its search-augmented variants reach 4.6--4.8\%. \texttt{openai-deepresearch} has 3.5\% hallucinated but 10.1\% non-resolving. The Claude models achieve the lowest hallucination rates (3.0--3.2\%) with smaller samples (641--1,735 URLs), yielding wider CIs. Search-augmented GPT models range from 5.4\% to 8.8\% hallucinated, though their small sample sizes (296--402 URLs) produce CIs spanning 3--5 percentage points.

\begin{table}[t]
\centering
\resizebox{\columnwidth}{!}{%
\begin{tabular}{llrrrr}
\toprule
\textbf{Model} & \textbf{Provider} & \textbf{URLs} & \textbf{\% Non-resolving} & \textbf{\% Hallucinated} & \textbf{\% Stale} \\
\midrule
\texttt{claude-3-5-sonnet-search} & Anthropic & 641 & 7.8\ci{5.8}{10.0} & 3.0\ci{1.7}{4.4} & 4.8 \\
\texttt{claude-3-7-sonnet-search} & Anthropic & 1,735 & 8.5\ci{7.1}{9.7} & 3.2\ci{2.4}{4.1} & 5.2 \\
\texttt{openai-deepresearch} & OpenAI & 4,121 & 10.1\ci{9.1}{11.0} & 3.5\ci{3.0}{4.1} & 6.6 \\
\texttt{gemini-2.5-flash-search} & Google & 2,433 & 5.4\ci{4.5}{6.3} & 4.6\ci{3.7}{5.4} & 0.8 \\
\texttt{gemini-2.5-pro-search} & Google & 1,609 & 5.9\ci{4.8}{7.1} & 4.8\ci{3.8}{5.9} & 1.1 \\
\texttt{gpt-4.1} & OpenAI & 336 & 5.4\ci{3.0}{7.7} & 5.4\ci{3.0}{7.7} & 0.0 \\
\texttt{gpt-4.1-mini} & OpenAI & 296 & 7.4\ci{4.4}{10.5} & 7.4\ci{4.7}{10.5} & 0.0 \\
\texttt{gpt-4o-search-preview} & OpenAI & 387 & 8.8\ci{5.9}{11.6} & 8.8\ci{6.2}{11.6} & 0.0 \\
\texttt{gpt-4o-mini-search-prev.} & OpenAI & 402 & 8.7\ci{6.0}{11.4} & 8.7\ci{6.0}{11.4} & 0.0 \\
\texttt{gemini-2.5-pro-deepres.} & Google & 11,309 & 18.5\ci{17.8}{19.2} & 13.3\ci{12.7}{13.9} & 5.2 \\
\bottomrule
\end{tabular}}%
\caption{DRBench per-model URL validity results sorted by hallucinated URL rate. Brackets show bootstrap 95\% CIs. ``\% Stale'' = \% Non-resolving $-$ \% Hallucinated. Model names abbreviated for space.}
\label{tab:drbench-results}
\end{table}

\paragraph{ExpertQA.} Our ExpertQA evaluation encompasses 168,021 URLs collected across 3 models and 32 academic fields, with an overall non-resolving URL rate of 8.22\% [8.09, 8.36]. \texttt{gpt-5.1} generates the most URLs per question (46.4 mean, 83,736 total) with an 8.47\% [8.28, 8.66] non-resolving rate.\footnote{\texttt{gpt-5.1} produces an unusually high proportion of Reddit URLs (18.2\% of all citations, compared to $<$1\% for other models). Because Reddit aggressively rate-limits automated URL checks, we classify Reddit URLs as alive based on manual spot-checks confirming they resolve individually. A sensitivity analysis treating all Reddit URLs as non-resolving raises GPT-5.1's rate to 26.7\% but does not affect the other models; see Appendix~\ref{app:sensitivity}.} \texttt{claude-sonnet-4-5} generates 28.3 URLs per question with a 9.38\% [9.16, 9.62] non-resolving rate (61,407 URLs). \texttt{gemini-2.5-pro} generates the fewest URLs per question (10.7 mean) and achieves the lowest non-resolving rate at 4.20\% [3.94, 4.47] (22,878 URLs).

\subsection{RQ2: How do deep research agents compare with search-augmented LLMs?}
\label{sec:rq2}

Deep research agents cite far more URLs per query than search-augmented models. \texttt{gemini-2.5-pro-deepresearch} produces 113.1 URLs per query, \texttt{openai-deepresearch} produces 41.2, and search-augmented models produce 3.0--24.3. The volume difference is substantial, but volume alone does not determine reliability.

Pooling across the two deep research agents, the hallucination rate is 10.7\% [10.2, 11.2] versus 4.8\% [4.3, 5.2] for the eight search-augmented models (two-proportion $z = 15.15$, $p < 10^{-51}$). Non-resolving rates show a similar gap: 16.2\% [15.7, 16.8] versus 6.8\% [6.2, 7.3] ($z = 20.20$, $p < 10^{-89}$). \texttt{gemini-2.5-pro-deepresearch} exhibits the highest hallucination rate (13.3\%) among all evaluated models, which suggests that multi-step retrieval and synthesis may amplify URL fabrication by compounding errors across retrieval steps. However, \texttt{openai-deepresearch} presents a contrasting case: despite its high URL volume, its hallucination rate is only 3.5\%, though its overall non-resolving URL rate reaches 10.1\%. Most of its non-resolving URLs are stale rather than fabricated.

We hypothesize that multi-step synthesis may produce URLs by blending patterns from retrieved pages, yielding plausible but nonexistent addresses. The lower hallucination rate of \texttt{openai-deepresearch} may reflect tighter coupling between retrieval and generation---for example, restricting URLs to those actually visited during browsing. We cannot verify this without access to internal architectures.

\subsection{RQ3: How strongly does citation reliability vary across academic fields?}
\label{sec:rq3}

\begin{figure}[t]
\centering
\includegraphics[width=\linewidth]{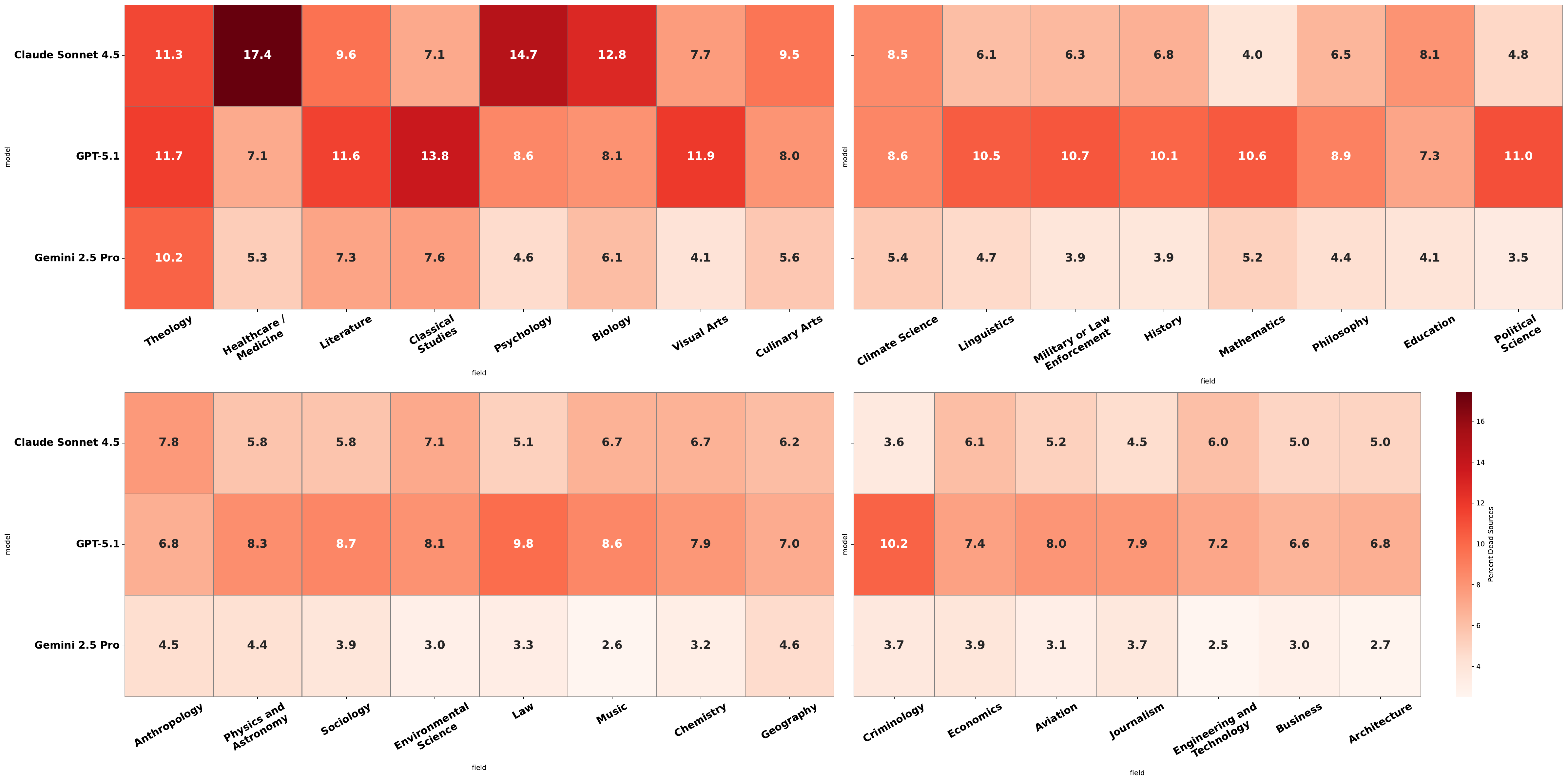}
\caption{Non-resolving URL rates by academic field and model for ExpertQA. Fields are sorted by overall non-resolving URL rate.}
\label{fig:expertqa-heatmap}
\end{figure}

Figure~\ref{fig:expertqa-heatmap} reports non-resolving URL rates stratified by academic field. Rates range from 5.4\% [4.9, 5.9] (Business) to 11.4\% [8.1, 14.6] (Theology), a roughly 2$\times$ variation (two-proportion $z = 4.83$, $p < 10^{-5}$). Fields with fast-changing or specialized web content---Theology, Classical Studies, Healthcare/Medicine---cluster at the top; fields with stable, mainstream web presence---Business, Architecture, Journalism---cluster at the bottom.

Domain effects are more pronounced per-model than in aggregate. Claude Sonnet~4.5 exhibits the widest field variation: 4.0\% [2.7, 5.3] (Mathematics) to 17.4\% [16.8, 18.1] (Healthcare/Medicine), a 4.3$\times$ difference ($z = 10.34$, $p < 10^{-24}$). Gemini consistently achieves the lowest rates across nearly all fields (2.5\%--10.2\%), suggesting its retrieval pipeline is more robust to domain-level variation. Healthcare/Medicine is consistently problematic across all three models (17.4\% [16.8, 18.1] Claude, 7.1\% [6.6, 7.5] GPT, 5.3\% [4.8, 5.9] Gemini)---precisely the domain where citation errors risk propagating into clinical recommendations.

Detailed field-level tables, contributing factors, and subfield-level analysis (where within-field variation can exceed between-field variation) are reported in Appendix~\ref{app:domain-analysis}.

\subsection{RQ4: What fraction of citation failures is fabrication versus link rot?}
\label{sec:rq4}

The gap between non-resolving and hallucinated URL rates varies substantially across models and is diagnostically informative, because it separates two failure modes that require different mitigations.

Four OpenAI search-augmented models (\texttt{gpt-4.1}, \texttt{gpt-4.1-mini}, \texttt{gpt-4o-search-preview}, \texttt{gpt-4o-mini-search-preview}) show non-resolving rates equal to hallucinated rates, with zero stale URLs. When these models cite a non-resolving URL, it is always fabricated. This pattern is consistent with URL generation that is not grounded in actual retrieval results, though we cannot confirm this without access to the internal retrieval pipeline.

\texttt{openai-deepresearch}, the two Claude models, and \texttt{gemini-2.5-pro-deepresearch} all show a substantial gap between non-resolving and hallucinated rates. \texttt{openai-deepresearch} has 10.1\% non-resolving but only 3.5\% hallucinated, meaning 65\% of its non-resolving URLs are stale---real pages that went offline. The Claude models show 4.8\% and 5.2\% stale URL rates, a pattern consistent with genuine web retrieval that occasionally returns stale index entries. We interpret these gaps as evidence of genuine web retrieval operating over a partially stale index.

The stale-vs-hallucinated distinction has direct implications for mitigation strategy. Hallucinated URLs require better URL generation or grounding, whereas stale URLs require fresher search indices or post-generation URL validation. Section~\ref{sec:urlhealth} introduces \texttt{urlhealth}, which automates this stale-vs-hallucinated classification.

\subsection{RQ5: Does producing more citations improve per-citation reliability?}
\label{sec:rq5}

The relationship between citation volume and citation reliability is, if anything, inverse. \texttt{gpt-5.1} generates 46.4 URLs per question, roughly 4.3$\times$ more than \texttt{gemini-2.5-pro} (10.7), with a 2$\times$ higher non-resolving rate (8.5\% vs. 4.2\%). \texttt{claude-sonnet-4-5} falls between (28.3 URLs/question, 9.4\%). In the DRBench data, \texttt{gemini-2.5-pro-deepresearch} produces the most URLs per query (113.1) and also has the highest hallucination rate (13.3\%).

Generating more citations does not compensate for, and may exacerbate, per-citation quality problems. Models that retrieve more sources may be less selective, including marginal or poorly matched results that are more likely to fail verification. Citation volume is not a proxy for citation quality.

%% ============================================================
%% SECTION 5: MITIGATION
%% ============================================================
\section{Mitigation}
\label{sec:mitigation}
\label{sec:urlhealth}

\textbf{RQ6: Can post-hoc URL verification reduce citation hallucination?} Section~\ref{sec:characterization} finds URL-level citation failures at rates of 5--18\% non-resolving and 3--13\% hallucinated. We address this with \texttt{urlhealth}, an open-source, model-agnostic URL verification tool that operates on LLM output after generation. For a given URL, \texttt{urlhealth} issues an HTTP HEAD request (falling back to GET when necessary) and classifies the result into four categories: \texttt{LIVE} (HTTP~200), \texttt{DEAD} (HTTP~404 with a Wayback Machine snapshot, corresponding to stale URLs), \texttt{LIKELY\_HALLUCINATED} (HTTP~404 with no archived snapshot), or \texttt{UNKNOWN} (other status codes or connection failures requiring manual inspection). The library is 83 lines of Python, available as a pip-installable package and as an agentskills.io skill for integration with AI coding agents. Implementation details and comparison to the ad-hoc verification methodology are provided in Appendix~\ref{app:experimental}.

\subsection{Agentic self-correction evaluation}
\label{sec:mitigation-eval}

We evaluate \texttt{urlhealth} as a tool within an agentic self-correction loop. Each model receives 435 ExpertQA questions (a 20\% sample) and generates answers with cited URLs.\footnote{The 20\% sample size reflects practical constraints: the Anthropic experiment hit a monthly API usage limit after 658 questions. The first 435 questions common to all three runs were used for uniform comparison.} The model has access to \texttt{urlhealth} as a callable tool and can iteratively verify and correct its own citations across multiple rounds. Claude Sonnet~4.5 and GPT-5.1 invoke \texttt{urlhealth} as a function tool alongside web search within the same generation turn. Gemini~2.5~Pro uses a two-phase approach---Google Search grounding generates the initial response, then a separate turn invokes \texttt{urlhealth} for verification---because the Gemini API does not permit simultaneous use of grounding and custom function-calling tools. Table~\ref{tab:urlcheck-results} summarizes URL health outcomes and correction behavior across models.

\paragraph{Results.} All three models achieve near-zero confirmed-broken citations in their final responses. Using consistent \texttt{urlhealth} classification for both pre- and post-tool conditions on the same question subset, the non-resolving rate drops significantly for all three models (all $p < 10^{-35}$, two-proportion $z$-test): GPT-5.1 from 16.0\% to 0.6\% ($26\times$), Gemini from 6.1\% to 0.1\% ($79\times$), and Claude from 4.9\% to 0.8\% ($6.4\times$). Gemini~2.5~Pro reaches the highest live-URL rate (88.9\% [88.0, 89.9]) with 0.5\% [0.3, 0.8] hallucination. Claude Sonnet~4.5 follows at 79.3\% [78.4, 80.2] live with 0.4\% [0.3, 0.6] hallucination but a larger UNKNOWN fraction (20.2\% [19.3, 21.1]), reflecting timeouts, paywalls, and bot-blocking. GPT-5.1 achieves 78.0\% [76.8, 79.1] live with 1.8\% [1.4, 2.1] hallucination and 19.7\% [18.6, 20.8] UNKNOWN. All intervals are bootstrap 95\% CIs. The models differ in citation volume: Claude generates 18.4~URLs per question (14.6 live), GPT-5.1 generates 11.1 (8.7 live), and Gemini generates 9.7 (8.6 live). The distribution of correction rounds is shown in Appendix~\ref{app:self-correction}.

\paragraph{Tool-use competence matters.} In preliminary experiments with \texttt{gpt-5-nano}, the same pipeline produced a 7.5\% NOT\_LIVE rate in final responses versus 0.6\% for GPT-5.1, with 48 hallucinated URLs persisting across up to 14 correction rounds. The smaller model called the verification tool but failed to act on its results, repeatedly re-proposing flagged URLs. Replacing it with GPT-5.1 reduced the NOT\_LIVE rate to 0.6\%, matching Claude and Gemini. Tool-based mitigation thus requires not just tool \emph{access} but competent tool \emph{use}. The UNKNOWN category (10--20\% across models) represents a practical ceiling on automated verification: bot-blocking, paywalls, and ambiguous HTTP responses prevent definitive classification. A headless-browser audit of 600 stratified-sampled UNKNOWN URLs finds that 89.0\% [86.3, 91.5] are live or blocked-but-accessible, with only 11.0\% [8.5, 13.7] genuinely dead (Appendix~\ref{app:unknown-audit}).

\begin{table}[t]
\centering
\small
\caption{URL health outcomes and self-correction behavior for three models on 435 ExpertQA questions with \texttt{urlhealth} as an agentic tool. Percentages are relative to total URLs proposed by each model. Brackets show bootstrap 95\% CIs.}
\label{tab:urlcheck-results}
\begin{tabular}{@{}lrrr@{}}
\toprule
& \textbf{Claude} & \textbf{Gemini} & \textbf{GPT-5.1} \\
\midrule
\multicolumn{4}{@{}l}{\textit{URL health}} \\
Total URLs & 7{,}985 & 4{,}203 & 4{,}829 \\
URLs / question & 18.4 & 9.7 & 11.1 \\
LIVE (\%) & 79.3\ci{78.4}{80.2} & 88.9\ci{88.0}{89.9} & 78.0\ci{76.8}{79.1} \\
DEAD (\%) & 0.1\ci{0.1}{0.2} & 0.2\ci{0.0}{0.3} & 0.6\ci{0.4}{0.8} \\
LIKELY\_HALL. (\%) & 0.4\ci{0.3}{0.6} & 0.5\ci{0.3}{0.8} & 1.8\ci{1.4}{2.1} \\
UNKNOWN (\%) & 20.2\ci{19.3}{21.1} & 10.3\ci{9.4}{11.3} & 19.7\ci{18.6}{20.8} \\
Live URLs / question & 14.6 & 8.6 & 8.7 \\
\midrule
\multicolumn{4}{@{}l}{\textit{Correction rounds}} \\
Mean & 4.06 & 1.56 & 2.52 \\
Median & 4 & 2 & 2 \\
Max & 8 & 2 & 6 \\
\bottomrule
\end{tabular}
\end{table}

%% ============================================================
%% SECTION 6: RELATED WORK (condensed)
%% ============================================================
\section{Related work}
\label{sec:related}

Hallucination taxonomies distinguish factual fabrication from faithfulness failures \citep{ji2023hallucination,huang2024hallucination}, and detection methods range from self-consistency probes \citep{manakul2023selfcheckgpt} to atomic-fact decomposition \citep{min2023factscore}. A parallel line of work on attribution asks whether generated text is verifiable against cited sources. \citet{rashkin2023ais} defined Attributable to Identified Sources (AIS), noting non-resolving links as an edge case; \citet{gao2023alce} benchmarked inline citation quality and found that even the best models lack complete citation support 50\% of the time; and \citet{liu2023verifiability} audited generative search engines, finding only 51.5\% of sentences fully supported by their citations. Retrieval-augmented generation \citep{lewis2020rag,nakano2021webgpt,menick2022gophercite} eliminates URL fabrication by construction when retrieval is from a known corpus, but production systems may generate URLs from parametric memory rather than actual browsing. \citet{wu2025sourcecheckup} verified medical citation URLs and found 40--70\% validity for API-only models, concluding that RAG-enabled models avoid URL hallucination---a conclusion our broader evaluation qualifies. The attribution literature asks ``does the source support the claim?'' Our work asks a logically prior question: ``does the source exist?''

Early empirical studies documented that LLMs fabricate scholarly citations wholesale \citep{alkaissi2023chatgpt}, with systematic rates of 55\% for \texttt{GPT-3.5} and 18\% for \texttt{GPT-4} \citep{walters2023fabrication}. Domain-specific investigations reveal substantial variation: DOI hallucination reaches 89.4\% in the humanities versus 29.1\% in natural sciences \citep{mugaanyi2024doi}, legal hallucination ranges from 58\% to 88\% depending on model and court hierarchy \citep{dahl2024legal}, and medical fabrication rates vary with topic prevalence \citep{linardon2025mental,chelli2024hallucination,aljamaan_reference_2024}. At larger scale, GhostCite \citep{xu2026ghostcite} benchmarks 13 LLMs across 40 domains (375K citations) with hallucination rates from 14\% to 95\%, while CiteAudit \citep{yuan2026citeaudit} achieves 97\% verification accuracy on generated benchmarks but only 90\% on real-world cases. \citet{algaba_large_2024,algaba_how_2025} show that LLM-generated references exhibit strong popularity bias, with existence rates of 40--50\% and 90\% of valid references falling among the top 10\% most-cited papers. The problem has reached peer-reviewed venues: \citet{sakai_hallucitation_2026} find a $>$10$\times$ increase in hallucinated references in ACL anthology papers from 2024 to 2025, and \citet{ansari_compound_2026} taxonomize fabrication modes in NeurIPS 2025 papers. Nearly all of this work studies non-retrieval settings where models generate references from parametric memory. Our work shows that citation hallucination persists even when models have access to web search, with 3--13\% of URLs fabricated in retrieval-augmented settings.

Deep research agents \citep{yao2023react,singh2025agenticrag,shao2024storm} iteratively search, read, and synthesize multi-page reports with citations. DRBench \citep{du2025drbench} evaluates citation accuracy (78--94\% across systems) but assumes cited URLs are retrievable. DRACO \citep{zhong2026draco} finds citation quality and factual accuracy are the weakest performance axes, with the best system achieving only 65\% citation quality. On the verification side, RARR \citep{gao2023rarr} repairs unsupported claims but operates at the claim level and assumes supporting evidence exists; SemanticCite \citep{haan_semanticcite__2025} classifies citation support with 84\% accuracy but requires the cited source to be accessible; and CiteGuard \citep{choi_citeguard__2026} approaches human-level citation attribution (68\% vs.\ 70\%) yet fails silently on fabricated URLs. Link rot compounds the problem: over 70\% of URLs in the Harvard Law Review no longer resolve \citep{zittrain2014perma}, and 25\% of webpages from 2013--2023 have disappeared \citep{pewresearch2024linkrot}. Existing benchmarks evaluate citation \emph{support} but not citation \emph{existence}. If 3--13\% of URLs are fabricated, support metrics systematically overestimate reliability. An extended discussion of related work is provided in Appendix~\ref{app:related}.

%% ============================================================
%% SECTION 7: CONCLUSION
%% ============================================================
\section{Conclusion}
\label{sec:conclusion}

We addressed six research questions about citation URL validity across commercial LLMs and deep research agents. \textbf{RQ1:} 3--13\% of citation URLs are hallucinated and 5--18\% are non-resolving across 10 models on DRBench and 3 on ExpertQA. \textbf{RQ2:} Deep research agents exhibit the highest hallucination rates; retrieval architecture matters more than output volume. \textbf{RQ3:} Non-resolving rates vary 2$\times$ across fields (5.4\%--11.4\%), with per-model variation reaching 4.3$\times$. \textbf{RQ4:} The stale-vs-hallucinated decomposition is diagnostically informative: some models fabricate every non-resolving URL, while others show stale fractions indicating genuine retrieval. \textbf{RQ5:} More citations per query do not mean fewer errors per citation. \textbf{RQ6:} \texttt{urlhealth} reduces non-resolving citation URLs by $6\textrm{--}79\times$ to under 1\% in agentic self-correction, though effectiveness depends on tool-use competence.

Two directions seem most pressing. First, generation-time constraints restricting URL emission to actually-visited pages could reduce fabrication at its source. Second, extending verification to bibliographic metadata and fabricated snippets would provide a more complete picture of citation reliability.

We are committed to releasing \texttt{urlhealth} and all experimental data and scripts pertaining to this work under MIT license to further LLM/agent driven bibliography verification and mitigation research.

\section*{Acknowledgments}
This research was developed with funding from the Defense Advanced Research Projects Agency's (DARPA) SciFy program (Agreement No. HR00112520300). The views expressed are those of the author and do not reflect the official policy or position of the Department of Defense or the U.S. Government.

%% ============================================================
%% LIMITATIONS (brief, after Conclusion)
%% ============================================================
\section*{Limitations}
\label{sec:limitations}
Our hallucination estimates rely on Wayback Machine coverage, which is substantial but incomplete; some hallucinated URLs may be misclassified as stale, and vice versa. URL liveness is a point-in-time measurement, subject to false positives from bot-blocking despite mitigations. A headless-browser audit of UNKNOWN URLs (Appendix~\ref{app:unknown-audit}) confirms that 89\% are live or blocked, supporting the lower-bound interpretation. We evaluate models from three providers on two English-dominated benchmarks; results may not generalize to all models or languages. A detailed discussion is in Appendix~\ref{app:limitations}.

%% ============================================================
%% ETHICS STATEMENT
%% ============================================================
\section*{Ethics Statement}

This study involves no human subjects. All experiments query commercial LLM APIs under their standard terms of service. URL liveness checks use HTTP HEAD requests with rate limiting, imposing negligible load on target servers. No private or personally identifiable data is collected: inputs are drawn from publicly available benchmarks (DRBench, ExpertQA) and outputs are model-generated URLs. We report per-model hallucination rates, which could be perceived negatively by specific vendors. We consider this reporting necessary for reproducibility and accountability, and note that all models tested are commercial products whose capabilities are already publicly advertised. We release \texttt{urlhealth} as an open-source diagnostic tool; it does not modify model outputs and carries no foreseeable dual-use risk.

\section*{Generative AI Use Disclosure}
\label{sec:generative-ai-use-disclosure}

The authors acknowledge the use of LLMs in the preparation of this manuscript. Specifically, the authors used Gemini-Pro 3.0 and Claude Sonnet 4.5 to proofread the drafts, make plots, and provide light rewrites. In all such cases, the corresponding author has manually reviewed the output for its accuracy.

%% ============================================================
%% ACKNOWLEDGMENTS
%% ============================================================
% \section*{Acknowledgments}

% Acknowledgments commented out for blind review

\bibliography{paper}

@misc{du2025drbench,
	title = {{DeepResearch} {Bench}: {A} {Comprehensive} {Benchmark} for {Deep} {Research} {Agents}},
	url = {http://arxiv.org/abs/2506.11763},
	doi = {10.48550/arXiv.2506.11763},
	publisher = {arXiv},
	author = {Du, Mingxuan and Xu, Benfeng and Zhu, Chiwei and Wang, Xiaorui and Mao, Zhendong},
	month = jun,
	year = {2025},
	note = {arXiv:2506.11763},
}

@misc{malaviya2024expertqa,
	title = {{ExpertQA}: {Expert}-{Curated} {Questions} and {Attributed} {Answers}},
	url = {http://arxiv.org/abs/2309.07852},
	doi = {10.48550/arXiv.2309.07852},
	publisher = {arXiv},
	author = {Malaviya, Chaitanya and Lee, Subin and Chen, Sihao and Sieber, Elizabeth and Yatskar, Mark and Roth, Dan},
	month = apr,
	year = {2024},
	note = {arXiv:2309.07852},
}

@article{ji2023hallucination,
	title = {Survey of {Hallucination} in {Natural} {Language} {Generation}},
	volume = {55},
	url = {https://dl.acm.org/doi/10.1145/3571730},
	doi = {10.1145/3571730},
	number = {12},
	journal = {ACM Computing Surveys},
	author = {Ji, Ziwei and Lee, Nayeon and Frieske, Rita and Yu, Tiezheng and Su, Dan and Xu, Yan and Ishii, Etsuko and Bang, Ye Jin and Madotto, Andrea and Fung, Pascale},
	month = dec,
	year = {2023},
	pages = {1--38},
}

@misc{huang2024hallucination,
	title = {A {Survey} on {Hallucination} in {Large} {Language} {Models}: {Principles}, {Taxonomy}, {Challenges}, and {Open} {Questions}},
	url = {http://arxiv.org/abs/2311.05232},
	doi = {10.48550/arXiv.2311.05232},
	publisher = {arXiv},
	author = {Huang, Lei and Yu, Weijiang and Ma, Weitao and Zhong, Weihong and Feng, Zhangyin and Wang, Haotian and Chen, Qianglong and Peng, Weihua and Feng, Xiaocheng and Qin, Bing and Liu, Ting},
	month = nov,
	year = {2024},
	note = {arXiv:2311.05232},
}

@misc{manakul2023selfcheckgpt,
	title = {{SelfCheckGPT}: {Zero}-{Resource} {Black}-{Box} {Hallucination} {Detection} for {Generative} {Large} {Language} {Models}},
	url = {http://arxiv.org/abs/2303.08896},
	doi = {10.48550/arXiv.2303.08896},
	publisher = {arXiv},
	author = {Manakul, Potsawee and Liusie, Adian and Gales, Mark J. F.},
	month = oct,
	year = {2023},
	note = {arXiv:2303.08896},
}

@misc{agrawal2024hallucinating,
	title = {Do {Language} {Models} {Know} {When} {They}'re {Hallucinating} {References}?},
	url = {http://arxiv.org/abs/2305.18248},
	doi = {10.48550/arXiv.2305.18248},
	publisher = {arXiv},
	author = {Agrawal, Ayush and Suzgun, Mirac and Mackey, Lester and Kalai, Adam Tauman},
	month = mar,
	year = {2024},
	note = {arXiv:2305.18248},
}

@misc{min2023factscore,
	title = {{FActScore}: {Fine}-grained {Atomic} {Evaluation} of {Factual} {Precision} in {Long} {Form} {Text} {Generation}},
	url = {http://arxiv.org/abs/2305.14251},
	doi = {10.48550/arXiv.2305.14251},
	publisher = {arXiv},
	author = {Min, Sewon and Krishna, Kalpesh and Lyu, Xinxi and Lewis, Mike and Yih, Wen-tau and Koh, Pang Wei and Iyyer, Mohit and Zettlemoyer, Luke and Hajishirzi, Hannaneh},
	month = oct,
	year = {2023},
	note = {arXiv:2305.14251},
}

@misc{lewis2020rag,
	title = {Retrieval-{Augmented} {Generation} for {Knowledge}-{Intensive} {NLP} {Tasks}},
	url = {http://arxiv.org/abs/2005.11401},
	doi = {10.48550/arXiv.2005.11401},
	publisher = {arXiv},
	author = {Lewis, Patrick and Perez, Ethan and Piktus, Aleksandra and Petroni, Fabio and Karpukhin, Vladimir and Goyal, Naman and K{\"u}ttler, Heinrich and Lewis, Mike and Yih, Wen-tau and Rockt{\"a}schel, Tim and Riedel, Sebastian and Kiela, Douwe},
	month = apr,
	year = {2021},
	note = {arXiv:2005.11401},
}

@misc{nakano2021webgpt,
	title = {{WebGPT}: {Browser}-assisted question-answering with human feedback},
	url = {http://arxiv.org/abs/2112.09332},
	doi = {10.48550/arXiv.2112.09332},
	publisher = {arXiv},
	author = {Nakano, Reiichiro and Hilton, Jacob and Balaji, Suchir and Wu, Jeff and Ouyang, Long and Kim, Christina and Hesse, Christopher and Jain, Shantanu and Kosaraju, Vineet and Saunders, William and Jiang, Xu and Cobbe, Karl and Eloundou, Tyna and Krueger, Gretchen and Button, Kevin and Knight, Matthew and Chess, Benjamin and Schulman, John},
	month = jun,
	year = {2022},
	note = {arXiv:2112.09332},
}

@misc{menick2022gophercite,
	title = {Teaching language models to support answers with verified quotes},
	url = {http://arxiv.org/abs/2203.11147},
	doi = {10.48550/arXiv.2203.11147},
	publisher = {arXiv},
	author = {Menick, Jacob and Trebacz, Maja and Mikulik, Vladimir and Aslanides, John and Song, Francis and Chadwick, Martin and Glaese, Mia and Young, Susannah and Campbell-Gillingham, Lucy and Irving, Geoffrey and McAleese, Nat},
	month = mar,
	year = {2022},
	note = {arXiv:2203.11147},
}

@article{rashkin2023ais,
	title = {Measuring {Attribution} in {Natural} {Language} {Generation} {Models}},
	volume = {49},
	url = {https://direct.mit.edu/coli/article/49/4/777/116438},
	doi = {10.1162/coli_a_00486},
	number = {4},
	journal = {Computational Linguistics},
	author = {Rashkin, Hannah and Nikolaev, Vitaly and Lamm, Matthew and Aroyo, Lora and Collins, Michael and Das, Dipanjan and Petrov, Slav and Tomar, Gaurav Singh and Turc, Iulia and Reitter, David},
	month = dec,
	year = {2023},
	pages = {777--840},
}

@misc{bohnet2023attributed,
	title = {Attributed {Question} {Answering}: {Evaluation} and {Modeling} for {Attributed} {Large} {Language} {Models}},
	url = {http://arxiv.org/abs/2212.08037},
	doi = {10.48550/arXiv.2212.08037},
	publisher = {arXiv},
	author = {Bohnet, Bernd and Tran, Vinh Q. and Verga, Pat and Aharoni, Roee and Andor, Daniel and Soares, Livio Baldini and Ciaramita, Massimiliano and Eisenstein, Jacob and Ganchev, Kuzman and Herzig, Jonathan and Hui, Kai and Kwiatkowski, Tom and Ma, Ji and Ni, Jianmo and Saralegui, Lierni Sestorain and Schuster, Tal and Cohen, William W. and Collins, Michael and Das, Dipanjan and Metzler, Donald and Petrov, Slav and Webster, Kellie},
	month = feb,
	year = {2023},
	note = {arXiv:2212.08037},
}

@misc{gao2023rarr,
	title = {{RARR}: {Researching} and {Revising} {What} {Language} {Models} {Say}, {Using} {Language} {Models}},
	url = {http://arxiv.org/abs/2210.08726},
	doi = {10.48550/arXiv.2210.08726},
	publisher = {arXiv},
	author = {Gao, Luyu and Dai, Zhuyun and Pasupat, Panupong and Chen, Anthony and Chaganty, Arun Tejasvi and Fan, Yicheng and Zhao, Vincent Y. and Lao, Ni and Lee, Hongrae and Juan, Da-Cheng and Guu, Kelvin},
	month = may,
	year = {2023},
	note = {arXiv:2210.08726},
}

@misc{gao2023alce,
	title = {Enabling {Large} {Language} {Models} to {Generate} {Text} with {Citations}},
	url = {http://arxiv.org/abs/2305.14627},
	doi = {10.48550/arXiv.2305.14627},
	publisher = {arXiv},
	author = {Gao, Tianyu and Yen, Howard and Yu, Jiatong and Chen, Danqi},
	month = oct,
	year = {2023},
	note = {arXiv:2305.14627},
}

@inproceedings{liu2023verifiability,
	title = {Evaluating {Verifiability} in {Generative} {Search} {Engines}},
	url = {https://aclanthology.org/2023.findings-emnlp.467},
	doi = {10.18653/v1/2023.findings-emnlp.467},
	booktitle = {Findings of the {Association} for {Computational} {Linguistics}: {EMNLP} 2023},
	publisher = {Association for Computational Linguistics},
	author = {Liu, Nelson and Zhang, Tianyi and Liang, Percy},
	year = {2023},
	pages = {7001--7025},
}

@article{wu2025sourcecheckup,
	title = {An automated framework for assessing how well {LLMs} cite relevant medical references},
	volume = {16},
	url = {https://www.nature.com/articles/s41467-025-58551-6},
	doi = {10.1038/s41467-025-58551-6},
	number = {1},
	journal = {Nature Communications},
	author = {Wu, Kevin and Wu, Eric and Wei, Kevin and Zhang, Angela and Casasola, Allison and Nguyen, Teresa and Riantawan, Sith and Shi, Patricia and Ho, Daniel and Zou, James},
	month = apr,
	year = {2025},
	pages = {3615},
}

@article{alkaissi2023chatgpt,
	title = {Artificial {Hallucinations} in {ChatGPT}: {Implications} in {Scientific} {Writing}},
	url = {https://www.cureus.com/articles/138667-artificial-hallucinations-in-chatgpt-implications-in-scientific-writing},
	doi = {10.7759/cureus.35179},
	journal = {Cureus},
	author = {Alkaissi, Hussam and McFarlane, Samy I},
	month = feb,
	year = {2023},
}

@article{walters2023fabrication,
	title = {Fabrication and errors in the bibliographic citations generated by {ChatGPT}},
	volume = {13},
	url = {https://www.nature.com/articles/s41598-023-41032-5},
	doi = {10.1038/s41598-023-41032-5},
	number = {1},
	journal = {Scientific Reports},
	author = {Walters, William H. and Wilder, Esther Isabelle},
	month = sep,
	year = {2023},
	pages = {14045},
}

@article{mugaanyi2024doi,
	title = {Evaluation of {Large} {Language} {Model} {Performance} and {Reliability} for {Citations} and {References} in {Scholarly} {Writing}: {Cross}-{Disciplinary} {Study}},
	volume = {26},
	url = {https://www.jmir.org/2024/1/e52935},
	doi = {10.2196/52935},
	number = {1},
	journal = {Journal of Medical Internet Research},
	author = {Mugaanyi, Joseph and Cai, Liuying and Cheng, Sumei and Lu, Caide and Huang, Jing},
	month = apr,
	year = {2024},
	pages = {e52935},
}

@misc{dahl2024legal,
	title = {Large {Legal} {Fictions}: {Profiling} {Legal} {Hallucinations} in {Large} {Language} {Models}},
	url = {http://arxiv.org/abs/2401.01301},
	doi = {10.48550/arXiv.2401.01301},
	publisher = {arXiv},
	author = {Dahl, Matthew and Magesh, Varun and Suzgun, Mirac and Ho, Daniel E.},
	month = jun,
	year = {2024},
	note = {arXiv:2401.01301},
}

@article{linardon2025mental,
	title = {Influence of {Topic} {Familiarity} and {Prompt} {Specificity} on {Citation} {Fabrication} in {Mental} {Health} {Research} {Using} {Large} {Language} {Models}: {Experimental} {Study}},
	volume = {12},
	url = {https://mental.jmir.org/2025/1/e80371},
	doi = {10.2196/80371},
	number = {1},
	journal = {JMIR Mental Health},
	author = {Linardon, Jake and Jarman, Hannah K. and McClure, Zoe and Anderson, Cleo and Liu, Claudia and Messer, Mariel},
	month = nov,
	year = {2025},
	pages = {e80371},
}

@article{chelli2024hallucination,
	title = {Hallucination {Rates} and {Reference} {Accuracy} of {ChatGPT} and {Bard} for {Systematic} {Reviews}: {Comparative} {Analysis}},
	volume = {26},
	url = {https://www.jmir.org/2024/1/e53164},
	doi = {10.2196/53164},
	journal = {Journal of Medical Internet Research},
	author = {Chelli, Mika{\"e}l and Descamps, Jules and Lavou{\'e}, Vincent and Trojani, Christophe and Azar, Michel and Deckert, Marcel and Raynier, Jean-Luc and Clowez, Gilles and Boileau, Pascal and Ruetsch-Chelli, Caroline},
	month = may,
	year = {2024},
	pages = {e53164},
}

@misc{xu2026ghostcite,
	title = {{GhostCite}: {A} {Large}-{Scale} {Analysis} of {Citation} {Validity} in the {Age} of {Large} {Language} {Models}},
	url = {http://arxiv.org/abs/2602.06718},
	doi = {10.48550/arXiv.2602.06718},
	publisher = {arXiv},
	author = {Xu, Zuyao and Qiu, Yuqi and Sun, Lu and Miao, FaSheng and Wu, Fubin and Wang, Xinyi and Li, Xiang and Lu, Haozhe and Zhang, ZhengZe and Hu, Yuxin and Li, Jialu and Luo, Jin and Zhang, Feng and Luo, Rui and Liu, Xinran and Li, Yingxian and Liu, Jiaji},
	month = feb,
	year = {2026},
	note = {arXiv:2602.06718},
}

@misc{yuan2026citeaudit,
	title = {{CiteAudit}: {You} {Cited} {It}, {But} {Did} {You} {Read} {It}? {A} {Benchmark} for {Verifying} {Scientific} {References} in the {LLM} {Era}},
	url = {http://arxiv.org/abs/2602.23452},
	doi = {10.48550/arXiv.2602.23452},
	publisher = {arXiv},
	author = {Yuan, Zhengqing and Shi, Kaiwen and Zhang, Zheyuan and Sun, Lichao and Chawla, Nitesh V. and Ye, Yanfang},
	month = feb,
	year = {2026},
	note = {arXiv:2602.23452},
}

@article{zittrain2014perma,
	title = {Perma: {Scoping} and {Addressing} the {Problem} of {Link} and {Reference} {Rot} in {Legal} {Citations}},
	url = {https://harvardlawreview.org/forum/vol-127/perma-scoping-and-addressing-the-problem-of-link-and-reference-rot-in-legal-citations/},
	doi = {10.2139/ssrn.2329161},
	journal = {Harvard Law Review Forum},
	author = {Zittrain, Jonathan and Albert, Kendra and Lessig, Lawrence},
	volume = {127},
	pages = {176--199},
	year = {2014},
}

@misc{pewresearch2024linkrot,
	title = {When {Online} {Content} {Disappears}},
	url = {https://www.pewresearch.org/data-labs/2024/05/17/when-online-content-disappears/},
	author = {{Pew Research Center}},
	month = may,
	year = {2024},
	note = {Accessed: 2026-03-02},
}

@misc{yao2023react,
	title = {{ReAct}: {Synergizing} {Reasoning} and {Acting} in {Language} {Models}},
	url = {http://arxiv.org/abs/2210.03629},
	doi = {10.48550/arXiv.2210.03629},
	publisher = {arXiv},
	author = {Yao, Shunyu and Zhao, Jeffrey and Yu, Dian and Du, Nan and Shafran, Izhak and Narasimhan, Karthik and Cao, Yuan},
	month = mar,
	year = {2023},
	note = {arXiv:2210.03629},
}

@misc{singh2025agenticrag,
	title = {Agentic {Retrieval}-{Augmented} {Generation}: {A} {Survey} on {Agentic} {RAG}},
	url = {http://arxiv.org/abs/2501.09136},
	doi = {10.48550/arXiv.2501.09136},
	publisher = {arXiv},
	author = {Singh, Aditi and Ehtesham, Abul and Kumar, Saket and Khoei, Tala Talaei},
	month = feb,
	year = {2025},
	note = {arXiv:2501.09136},
}

@misc{shao2024storm,
	title = {Assisting in {Writing} {Wikipedia}-like {Articles} {From} {Scratch} with {Large} {Language} {Models}},
	url = {http://arxiv.org/abs/2402.14207},
	doi = {10.48550/arXiv.2402.14207},
	publisher = {arXiv},
	author = {Shao, Yijia and Jiang, Yucheng and Kanell, Theodore A. and Xu, Peter and Khattab, Omar and Lam, Monica S.},
	month = apr,
	year = {2024},
	note = {arXiv:2402.14207},
}

@misc{zhong2026draco,
	title = {{DRACO}: a {Cross}-{Domain} {Benchmark} for {Deep} {Research} {Accuracy}, {Completeness}, and {Objectivity}},
	url = {http://arxiv.org/abs/2602.11685},
	doi = {10.48550/arXiv.2602.11685},
	publisher = {arXiv},
	author = {Zhong, Joey and Zhang, Hao and Southern, Clare and Yang, Jeremy and Wang, Thomas and Jung, Kate and Zhang, Shu and Yarats, Denis and Ho, Johnny and Ma, Jerry},
	month = feb,
	year = {2026},
	note = {arXiv:2602.11685},
}

@misc{press_citeme__2024,
	title = {{CiteME}: {Can} {Language} {Models} {Accurately} {Cite} {Scientific} {Claims}?},
	shorttitle = {{CiteME}},
	url = {http://arxiv.org/abs/2407.12861},
	doi = {10.48550/arXiv.2407.12861},
	urldate = {2026-03-04},
	publisher = {arXiv},
	author = {Press, Ori and Hochlehnert, Andreas and Prabhu, Ameya and Udandarao, Vishaal and Press, Ofir and Bethge, Matthias},
	month = nov,
	year = {2024},
	note = {arXiv:2407.12861},
}

@misc{sakai_hallucitation_2026,
	title = {{HalluCitation} {Matters}: {Revealing} the {Impact} of {Hallucinated} {References} with 300 {Hallucinated} {Papers} in {ACL} {Conferences}},
	shorttitle = {{HalluCitation} {Matters}},
	url = {http://arxiv.org/abs/2601.18724},
	doi = {10.48550/arXiv.2601.18724},
	urldate = {2026-03-04},
	publisher = {arXiv},
	author = {Sakai, Yusuke and Kamigaito, Hidetaka and Watanabe, Taro},
	month = jan,
	year = {2026},
	note = {arXiv:2601.18724},
}

@misc{xu_citeeval__2025,
	title = {{CiteEval}: {Principle}-{Driven} {Citation} {Evaluation} for {Source} {Attribution}},
	shorttitle = {{CiteEval}},
	url = {http://arxiv.org/abs/2506.01829},
	doi = {10.48550/arXiv.2506.01829},
	urldate = {2026-03-04},
	publisher = {arXiv},
	author = {Xu, Yumo and Qi, Peng and Chen, Jifan and Liu, Kunlun and Han, Rujun and Liu, Lan and Min, Bonan and Castelli, Vittorio and Gupta, Arshit and Wang, Zhiguo},
	month = jun,
	year = {2025},
	note = {arXiv:2506.01829},
}

@misc{ansari_compound_2026,
	title = {Compound {Deception} in {Elite} {Peer} {Review}: {A} {Failure} {Mode} {Taxonomy} of 100 {Fabricated} {Citations} at {NeurIPS} 2025},
	shorttitle = {Compound {Deception} in {Elite} {Peer} {Review}},
	url = {http://arxiv.org/abs/2602.05930},
	doi = {10.48550/arXiv.2602.05930},
	urldate = {2026-03-04},
	publisher = {arXiv},
	author = {Ansari, Samar},
	month = feb,
	year = {2026},
	note = {arXiv:2602.05930},
}

@misc{algaba_how_2025,
	title = {How {Deep} {Do} {Large} {Language} {Models} {Internalize} {Scientific} {Literature} and {Citation} {Practices}?},
	url = {http://arxiv.org/abs/2504.02767},
	doi = {10.48550/arXiv.2504.02767},
	urldate = {2026-03-04},
	publisher = {arXiv},
	author = {Algaba, Andres and Holst, Vincent and Tori, Floriano and Mobini, Melika and Verbeken, Brecht and Wenmackers, Sylvia and Ginis, Vincent},
	month = apr,
	year = {2025},
	note = {arXiv:2504.02767},
}

@misc{algaba_large_2024,
	title = {Large {Language} {Models} {Reflect} {Human} {Citation} {Patterns} with a {Heightened} {Citation} {Bias}},
	url = {http://arxiv.org/abs/2405.15739},
	doi = {10.48550/arXiv.2405.15739},
	urldate = {2026-03-04},
	publisher = {arXiv},
	author = {Algaba, Andres and Mazijn, Carmen and Holst, Vincent and Tori, Floriano and Wenmackers, Sylvia and Ginis, Vincent},
	month = aug,
	year = {2024},
	note = {arXiv:2405.15739},
}

@misc{li_bibagent__2026,
	title = {{BibAgent}: {An} {Agentic} {Framework} for {Traceable} {Miscitation} {Detection} in {Scientific} {Literature}},
	shorttitle = {{BibAgent}},
	url = {http://arxiv.org/abs/2601.16993},
	doi = {10.48550/arXiv.2601.16993},
	urldate = {2026-03-04},
	publisher = {arXiv},
	author = {Li, Peiran and Lin, Fangzhou and Xing, Shuo and Zheng, Xiang and Hong, Xi and Yang, Siyuan and Sun, Jiashuo and Tu, Zhengzhong and Ni, Chaoqun},
	month = jan,
	year = {2026},
	note = {arXiv:2601.16993},
}

@misc{haan_semanticcite__2025,
	title = {{SemanticCite}: {Citation} {Verification} with {AI}-{Powered} {Full}-{Text} {Analysis} and {Evidence}-{Based} {Reasoning}},
	shorttitle = {{SemanticCite}},
	url = {http://arxiv.org/abs/2511.16198},
	doi = {10.48550/arXiv.2511.16198},
	urldate = {2026-03-04},
	publisher = {arXiv},
	author = {Haan, Sebastian},
	month = nov,
	year = {2025},
	note = {arXiv:2511.16198},
}

@misc{choi_citeguard__2026,
	title = {{CiteGuard}: {Faithful} {Citation} {Attribution} for {LLMs} via {Retrieval}-{Augmented} {Validation}},
	shorttitle = {{CiteGuard}},
	url = {http://arxiv.org/abs/2510.17853},
	doi = {10.48550/arXiv.2510.17853},
	urldate = {2026-03-04},
	publisher = {arXiv},
	author = {Choi, Yee Man and Guo, Xuehang and Fung, Yi R. and Wang, Qingyun},
	month = jan,
	year = {2026},
	note = {arXiv:2510.17853},
}

@misc{abbonato_checkifexist__2026,
	title = {{CheckIfExist}: {Detecting} {Citation} {Hallucinations} in the {Era} of {AI}-{Generated} {Content}},
	shorttitle = {{CheckIfExist}},
	url = {http://arxiv.org/abs/2602.15871},
	doi = {10.48550/arXiv.2602.15871},
	urldate = {2026-03-04},
	publisher = {arXiv},
	author = {Abbonato, Diletta},
	month = jan,
	year = {2026},
	note = {arXiv:2602.15871},
}

@misc{chern_factool__2023,
	title = {{FacTool}: {Factuality} {Detection} in {Generative} {AI} -- {A} {Tool} {Augmented} {Framework} for {Multi}-{Task} and {Multi}-{Domain} {Scenarios}},
	shorttitle = {{FacTool}},
	url = {http://arxiv.org/abs/2307.13528},
	doi = {10.48550/arXiv.2307.13528},
	urldate = {2026-03-04},
	publisher = {arXiv},
	author = {Chern, I.-Chun and Chern, Steffi and Chen, Shiqi and Yuan, Weizhe and Feng, Kehua and Zhou, Chunting and He, Junxian and Neubig, Graham and Liu, Pengfei},
	month = jul,
	year = {2023},
	note = {arXiv:2307.13528},
}

@misc{zhang_longcite__2024,
	title = {{LongCite}: {Enabling} {LLMs} to {Generate} {Fine}-grained {Citations} in {Long}-context {QA}},
	shorttitle = {{LongCite}},
	url = {http://arxiv.org/abs/2409.02897},
	doi = {10.48550/arXiv.2409.02897},
	urldate = {2026-03-04},
	publisher = {arXiv},
	author = {Zhang, Jiajie and Bai, Yushi and Lv, Xin and Gu, Wanjun and Liu, Danqing and Zou, Minhao and Cao, Shulin and Hou, Lei and Dong, Yuxiao and Feng, Ling and Li, Juanzi},
	month = sep,
	year = {2024},
	note = {arXiv:2409.02897},
}

@misc{saxena_generation_time_2025,
	title = {Generation-{Time} vs. {Post}-hoc {Citation}: {A} {Holistic} {Evaluation} of {LLM} {Attribution}},
	shorttitle = {Generation-{Time} vs. {Post}-hoc {Citation}},
	url = {http://arxiv.org/abs/2509.21557},
	doi = {10.48550/arXiv.2509.21557},
	urldate = {2026-03-04},
	publisher = {arXiv},
	author = {Saxena, Yash and Bommireddy, Raviteja and Padia, Ankur and Gaur, Manas},
	month = dec,
	year = {2025},
	note = {arXiv:2509.21557},
}

@misc{ravichander_halogen__2025,
	title = {{HALoGEN}: {Fantastic} {LLM} {Hallucinations} and {Where} to {Find} {Them}},
	shorttitle = {{HALoGEN}},
	url = {http://arxiv.org/abs/2501.08292},
	doi = {10.48550/arXiv.2501.08292},
	urldate = {2026-03-04},
	publisher = {arXiv},
	author = {Ravichander, Abhilasha and Ghela, Shrusti and Wadden, David and Choi, Yejin},
	month = jan,
	year = {2025},
	note = {arXiv:2501.08292},
}

@article{aljamaan_reference_2024,
	title = {Reference {Hallucination} {Score} for {Medical} {Artificial} {Intelligence} {Chatbots}: {Development} and {Usability} {Study}},
	volume = {12},
	issn = {2291-9694},
	shorttitle = {Reference {Hallucination} {Score} for {Medical} {Artificial} {Intelligence} {Chatbots}},
	url = {https://medinform.jmir.org/2024/1/e54345},
	doi = {10.2196/54345},
	language = {en},
	urldate = {2026-03-04},
	journal = {JMIR Medical Informatics},
	author = {Aljamaan, Fadi and Temsah, Mohamad-Hani and Altamimi, Ibraheem and Al-Eyadhy, Ayman and Jamal, Amr and Alhasan, Khalid and Mesallam, Tamer A and Farahat, Mohamed and Malki, Khalid H},
	month = jul,
	year = {2024},
	pages = {e54345},
}

@inproceedings{ram_citation_2025,
	title = {Citation Drift: Measuring Reference Stability in Multi-Turn {LLM} Conversations},
	author = {Ram, Gokul Srinath Seetha},
	booktitle = {Proceedings of the Third Workshop for Artificial Intelligence for Scientific Publications},
	month = dec,
	year = {2025},
	address = {Mumbai, India and virtual},
	publisher = {Association for Computational Linguistics},
	url = {https://aclanthology.org/2025.wasp-main.20/},
	pages = {186--191},
}

@inproceedings{aly_learning_2024,
	title = {Learning to {Generate} {Answers} with {Citations} via {Factual} {Consistency} {Models}},
	url = {https://aclanthology.org/2024.acl-long.641},
	doi = {10.18653/v1/2024.acl-long.641},
	language = {en},
	urldate = {2026-03-04},
	booktitle = {Proceedings of the 62nd {Annual} {Meeting} of the {Association} for {Computational} {Linguistics} ({Volume} 1: {Long} {Papers})},
	publisher = {Association for Computational Linguistics},
	author = {Aly, Rami and Tang, Zhiqiang and Tan, Samson and Karypis, George},
	address = {Bangkok, Thailand},
	year = {2024},
	pages = {11876--11896},
}

@inproceedings{ye_effective_2024,
	title = {Effective {Large} {Language} {Model} {Adaptation} for {Improved} {Grounding} and {Citation} {Generation}},
	url = {https://aclanthology.org/2024.naacl-long.346},
	doi = {10.18653/v1/2024.naacl-long.346},
	language = {en},
	urldate = {2026-03-04},
	booktitle = {Proceedings of the 2024 {Conference} of the {North} {American} {Chapter} of the {Association} for {Computational} {Linguistics}: {Human} {Language} {Technologies} ({Volume} 1: {Long} {Papers})},
	publisher = {Association for Computational Linguistics},
	author = {Ye, Xi and Sun, Ruoxi and Arik, Sercan and Pfister, Tomas},
	address = {Mexico City, Mexico},
	year = {2024},
	pages = {6237--6251},
}

@misc{wallat_correctness_2024,
	title = {Correctness is not {Faithfulness} in {RAG} {Attributions}},
	url = {http://arxiv.org/abs/2412.18004},
	doi = {10.48550/arXiv.2412.18004},
	urldate = {2026-03-04},
	publisher = {arXiv},
	author = {Wallat, Jonas and Heuss, Maria and Rijke, Maarten de and Anand, Avishek},
	month = dec,
	year = {2024},
	note = {arXiv:2412.18004},
}

@misc{ainsworth_how_2013,
	title = {How {Much} of the {Web} {Is} {Archived}?},
	url = {http://arxiv.org/abs/1212.6177},
	doi = {10.48550/arXiv.1212.6177},
	urldate = {2026-03-08},
	publisher = {arXiv},
	author = {Ainsworth, Scott G. and AlSum, Ahmed and SalahEldeen, Hany and Weigle, Michele C. and Nelson, Michael L.},
	month = jan,
	year = {2013},
	note = {arXiv:1212.6177},
}

@misc{alsum_profiling_2013,
	title = {Profiling {Web} {Archive} {Coverage} for {Top}-{Level} {Domain} and {Content} {Language}},
	url = {http://arxiv.org/abs/1309.4008},
	doi = {10.48550/arXiv.1309.4008},
	urldate = {2026-03-08},
	publisher = {arXiv},
	author = {AlSum, Ahmed and Weigle, Michele C. and Nelson, Michael L. and Sompel, Herbert Van de},
	month = sep,
	year = {2013},
	note = {arXiv:1309.4008},
}

@inproceedings{alkwai_how_2015,
	address = {Knoxville Tennessee USA},
	title = {How {Well} {Are} {Arabic} {Websites} {Archived}?},
	isbn = {9781450335942},
	url = {https://dl.acm.org/doi/10.1145/2756406.2756912},
	doi = {10.1145/2756406.2756912},
	language = {en},
	urldate = {2026-03-08},
	booktitle = {Proceedings of the 15th {ACM}/{IEEE}-{CS} {Joint} {Conference} on {Digital} {Libraries}},
	publisher = {ACM},
	author = {Alkwai, Lulwah M. and Nelson, Michael L. and Weigle, Michele C.},
	month = jun,
	year = {2015},
	pages = {223--232},
}
\bibliographystyle{colm2026_conference}

\newpage
\appendix

%% ============================================================
%% APPENDIX A: EXPERIMENTAL DETAILS
%% ============================================================
\section{Experimental details}
\label{app:experimental}

\begin{table}[ht]
\centering
\begin{tabular}{lccc}
\toprule
\textbf{Dataset} & \textbf{Queries} & \textbf{Language} & \textbf{Use in this study} \\
\midrule
DRBench & 100 & ZH, EN & Cross-model URL validity \\
ExpertQA & 2,177 & EN & Domain-stratified analysis \\
\bottomrule
\end{tabular}
\caption{Datasets used in this study. DRBench provides pre-collected outputs from 23 deep research and search-augmented models, enabling cross-model comparison of citation URL validity. ExpertQA spans 32 academic and professional fields, enabling domain-stratified analysis of fabrication rates.}
\label{tab:datasets}
\end{table}

\paragraph{DRBench model exclusion criteria.} The DRBench dataset contains outputs from 23 models, of which we analyze 10 from three providers (Google, OpenAI, and Anthropic) for which reliable URL liveness data is available. We exclude the remaining 13 models for one of three reasons: no article text available (\texttt{kimi-researcher} returned only scoring metadata with no generated text or URLs), insufficient data (\texttt{claude-research} produced only 1 URL), or anomalous URL patterns that suggest no real web retrieval was performed (\texttt{doubao-deepresearch}, \texttt{gensee-search-gpt-5}, and \texttt{nvidia-aiq-research-assistant} all show 100\% hallucination rates with undefined non-resolving URL status). Full results for all models with extractable URLs are available in the supplementary materials.

\paragraph{ExpertQA API configurations.} We evaluate three models on ExpertQA: \texttt{claude-sonnet-4-5} (Anthropic, using the web\_search\_20250305 beta API), \texttt{gemini-2.5-pro} (Google, using the Google Search grounding tool), and \texttt{gpt-5.1} (OpenAI, using the Responses API with web\_search tool, search\_context\_size=``high''). Each model receives a system prompt instructing it to answer with citations from academic or scholarly sources and to return structured JSON containing a markdown answer and numbered citations with URLs.

\paragraph{URL extraction details.} For DRBench, we extract URLs from model-generated article text using regex matching of \texttt{https?://} patterns; the article texts were pre-collected by \citet{du2025drbench}. For ExpertQA, we extract URLs from provider-specific API response formats: OpenAI \texttt{web\_search\_call.action.sources}, Anthropic \texttt{web\_search\_tool\_result} content blocks, and Google grounding metadata.

\paragraph{Concurrency and pipeline details.} HTTP requests for URL liveness checking are executed concurrently using a thread pool with 100 workers for DRBench and 60 workers for ExpertQA. The User-Agent header is set to mimic a standard browser to reduce false positives from bot-blocking.

\paragraph{urlhealth comparison to ad-hoc methodology.} The ad-hoc methodology in the main experimental pipeline (Section~\ref{sec:url-extraction}) classifies all non-403 4xx/5xx responses as non-resolving and queries the Wayback Machine for each. \texttt{urlhealth} simplifies this by triggering Wayback Machine queries only on 404 responses and routing other error codes to \texttt{UNKNOWN}, trading recall for reduced false positives.

\paragraph{Sensitivity to classification decisions.}
\label{app:sensitivity}
Two classification decisions in our pipeline warrant sensitivity analysis: the exclusion of HTTP~403 responses and the treatment of Reddit URLs.

\textit{403 exclusion.} Across the three ExpertQA models, 403 responses account for 6.6\% (\texttt{gemini-2.5-pro}) to 17.0\% (\texttt{claude-sonnet-4-5}) of all URLs. These responses are concentrated among academic publisher domains: \texttt{sciencedirect.com} (${\sim}$1{,}157 total across models), \texttt{mdpi.com} (${\sim}$713), \texttt{researchgate.net} (${\sim}$630), \texttt{academic.oup.com} (${\sim}$397), \texttt{tandfonline.com} (${\sim}$215), \texttt{journals.sagepub.com} (${\sim}$116), \texttt{onlinelibrary.wiley.com} (${\sim}$91), and \texttt{pubs.acs.org} (${\sim}$90). The publisher concentration confirms that these 403 responses reflect bot-blocking rather than genuinely dead pages. The 403 rate varies by field---for example, Chemistry (10--22\%) and Architecture (21--25\%) show higher rates than Law (3--9\%)---driven by the prevalence of publisher-hosted sources in each field. Because these URLs are excluded from the non-resolving count, our reported non-resolving rates are lower bounds.

\textit{Reddit URL treatment.} \texttt{gpt-5.1} generates 15{,}273 Reddit URLs in the ExpertQA baseline (18.2\% of its total), all classified as alive via post-processing. The other models produce few or no Reddit URLs (0 for Claude, 166 for Gemini). Table~\ref{tab:sensitivity} reports non-resolving rates under three treatments: the baseline classification used in Section~\ref{sec:rq1}, exclusion of Reddit URLs entirely, and a worst-case scenario treating all Reddit URLs as non-resolving. Under the baseline, \texttt{gpt-5.1}'s non-resolving rate is 8.47\%; excluding Reddit URLs raises it to 10.36\%, reflecting the dilution effect of the Reddit URLs in the denominator. The worst case (26.71\%) is an upper bound, as manual spot-checks confirm that individual Reddit URLs resolve correctly. Claude and Gemini rates are unaffected or negligibly changed ($\leq$0.04 percentage points).

In DRBench, Reddit URLs are minimal (456 across all models, primarily from \texttt{gemini-2.5-pro-deepresearch}) relative to 53{,}090 total URLs, and do not materially affect the results.

Both classification decisions---excluding 403s and treating Reddit URLs as alive---are defensible given the evidence (publisher-dominated 403 domains, individually resolving Reddit URLs), but they should be understood as producing lower-bound estimates of non-resolving rates.

\begin{table}[ht]
\centering
\small
\caption{Sensitivity of ExpertQA non-resolving rates to Reddit URL classification. Baseline: Reddit URLs classified as alive (Section~\ref{sec:rq1}). Reddit-excluded: Reddit URLs removed from both numerator and denominator. Worst case: all Reddit URLs treated as non-resolving.}
\label{tab:sensitivity}
\begin{tabular}{@{}lccc@{}}
\toprule
\textbf{Scenario} & \textbf{Claude} & \textbf{Gemini} & \textbf{GPT-5.1} \\
\midrule
Baseline & 9.38\% & 4.20\% & 8.47\% \\
Reddit-excluded & 9.38\% & 4.24\% & 10.36\% \\
Reddit-as-non-resolving & 9.38\% & 4.93\% & 26.71\% \\
\midrule
Reddit URLs & 0 & 166 & 15{,}273 \\
Reddit \% of total & 0.0\% & 0.7\% & 18.2\% \\
\bottomrule
\end{tabular}
\end{table}

\paragraph{Headless-browser audit of UNKNOWN URLs.}
\label{app:unknown-audit}
The \texttt{urlhealth} tool classifies URLs with non-200, non-404 HTTP status codes (or connection failures) as \texttt{UNKNOWN}. In the ExpertQA baseline, 5{,}898 of 29{,}384 unique URLs (20.1\%) receive this label, dominated by HTTP~403 (56.3\%) and HTTP~429 (26.3\%), with connection errors (12.1\%) and other codes (5.3\%) accounting for the remainder. To estimate the true composition of this category, we drew a stratified random sample of 200 UNKNOWN URLs per model (600 total), proportionally allocated by HTTP status code, and re-probed each with a Playwright headless browser using a standard Chrome user-agent and JavaScript rendering.

Table~\ref{tab:unknown-audit} reports the results. Overall, 89.0\% [86.3, 91.5] of UNKNOWN URLs are not dead: 32.0\% render as fully live pages and 57.0\% show access-blocking behavior (paywalls, bot walls, rate-limit pages) from sites that are clearly operational. Only 11.0\% [8.5, 13.7] are genuinely dead. The breakdown by original HTTP status code is informative: 403 responses are 99.7\% not dead (confirming that our 403 exclusion correctly identifies bot-blocking), 429 responses are 99.1\% not dead (rate-limiting from live servers, predominantly Reddit), connection errors are mixed (52.3\% not dead, 47.7\% dead), and HTTP~202 responses are uniformly dead (async endpoints that never deliver content).

Per-model rates are consistent for Claude (91.5\% [87.5, 95.0] not dead) and Gemini (86.5\% [81.5, 91.0]), while GPT-5.1 shows 89.0\% [84.5, 93.0] not dead. Extrapolating to the full UNKNOWN population, approximately 5{,}249 of the 5{,}898 UNKNOWN URLs are live or blocked, and roughly 649 may be genuinely non-resolving. Because these potentially dead URLs are excluded from our reported non-resolving rates, the exclusion reinforces the lower-bound interpretation of all rates reported in this paper.

\begin{table}[ht]
\centering
\small
\caption{Headless-browser audit of 600 stratified-sampled UNKNOWN URLs from the ExpertQA baseline. Verdicts: \textit{true\_live} = page renders real content; \textit{blocked} = paywall, bot wall, or rate-limit page from an operational site; \textit{dead} = error page, empty content, or unreachable. Brackets show bootstrap 95\% CIs.}
\label{tab:unknown-audit}
\begin{tabular}{@{}lcccc@{}}
\toprule
& \textbf{True live} & \textbf{Blocked} & \textbf{Dead} & \textbf{Not dead} \\
\midrule
\multicolumn{5}{@{}l}{\textit{By model}} \\
\texttt{claude-sonnet-4-5} ($n$=200) & 19.5\% & 72.0\% & 8.5\% [5.0, 12.5] & 91.5\% [87.5, 95.0] \\
\texttt{gemini-2.5-pro} ($n$=200) & 22.0\% & 64.5\% & 13.5\% [9.0, 18.5] & 86.5\% [81.5, 91.0] \\
\texttt{gpt-5.1} ($n$=200) & 54.5\% & 34.5\% & 11.0\% [7.0, 15.5] & 89.0\% [84.5, 93.0] \\
\midrule
\multicolumn{5}{@{}l}{\textit{By original HTTP status code}} \\
403 ($n$=383) & 15.9\% & 83.8\% & 0.3\% & 99.7\% \\
429 ($n$=110) & 99.1\% & 0.0\% & 0.9\% & 99.1\% \\
Connection error ($n$=65) & 21.5\% & 30.8\% & 47.7\% & 52.3\% \\
Other ($n$=42) & 19.0\% & 4.8\% & 76.2\% & 23.8\% \\
\midrule
\textbf{Overall} ($n$=600) & 32.0\% & 57.0\% & 11.0\% [8.5, 13.7] & 89.0\% [86.3, 91.5] \\
\bottomrule
\end{tabular}
\end{table}

%% ============================================================
%% APPENDIX B: EXTENDED RELATED WORK
%% ============================================================
\section{Extended related work}
\label{app:related}

This appendix provides the full discussion of related work summarized in Section~\ref{sec:related}.

\subsection{Hallucination in LLMs}

\paragraph{Surveys and taxonomies.} \citet{ji2023hallucination} established the intrinsic/extrinsic hallucination distinction across NLG tasks and cataloged metrics and mitigation methods; their updated survey notes that hallucination is worse for long-tail and low-resource domains, a pattern our domain-stratified results confirm quantitatively. \citet{huang2024hallucination} refined the taxonomy for LLMs into \emph{factuality hallucination}---further split into factual contradiction and factual fabrication---and \emph{faithfulness hallucination}. Citation URL fabrication falls under factual fabrication, because the model generates unverifiable structured artifacts (URLs) rather than incorrect prose claims. Huang et al.\ also identify three knowledge-boundary issues that contribute to fabrication: long-tail knowledge (less popular entities), temporal cutoffs (recently published content), and restricted content. All three map to failure modes we observe.

\paragraph{Detection via self-consistency.} \citet{manakul2023selfcheckgpt} detect hallucinations with SelfCheckGPT by sampling multiple responses and measuring consistency, achieving AUC-PR above 93\% for sentence-level detection on \texttt{GPT-3} biographies. The method operates as a black-box probe---requiring only text outputs, the same constraint we face with commercial APIs---but targets prose content rather than structured citations. \citet{agrawal2024hallucinating} applied self-consistency specifically to hallucinated references, treating them as a ``model organism'' for hallucination research. Their study found hallucination rates of 47\% (\texttt{GPT-4}) to 77\% (\texttt{Llama~2~7B}) for computer science reference titles, and showed that LLMs produce inconsistent author lists for fabricated references while accurately recalling authors of real ones. Intuitively, this suggests that information sufficient to avoid hallucination exists in the model but is not leveraged during generation. \citet{ravichander_halogen__2025} benchmark hallucination across nine domains including scientific attribution (verified against Semantic Scholar), and their HALOGEN framework classifies errors by source: incorrect recollection from training data, corrupted training data, or fabrication where no relevant training data exists. Note that citation fabrication is not monolithic; the same model may fail differently depending on what its training corpus contained. \citet{min2023factscore} introduced FActScore, an atomic-fact decomposition for evaluating long-form generation, and found that factual precision drops from 80\% for frequent entities to 16\% for rare ones---an entity-rarity effect directly analogous to our domain-dependent URL fabrication rates. They also evaluated the search-augmented PerplexityAI and found that its citations ``have little correlation with factual precision,'' an early signal that citation quality deserves independent study.

These surveys and benchmarks target \emph{claim-level} factual errors. Our work targets a different failure mode: \emph{citation-level} fabrication, where the citations themselves are fabricated regardless of whether the underlying claim is true. URL existence is binary and externally verifiable, making it amenable to automated evaluation in a way that free-text factual claims are not.

\subsection{Retrieval-augmented generation and attribution}

\paragraph{RAG and search-augmented LLMs.} \citet{lewis2020rag} introduced retrieval-augmented generation, combining a pre-trained seq2seq model with a dense retriever over a fixed Wikipedia index. Because retrieval is from a known corpus, URL fabrication cannot occur by construction. \citet{nakano2021webgpt} extended this paradigm to open-web browsing: \texttt{WebGPT} fine-tunes \texttt{GPT-3} to search and navigate real web pages, collecting references from pages actually visited, so that URL fabrication again cannot occur. The paper notes, however, that models are ``not trained to judge the reliability of sources, only to use them as evidence.'' \citet{menick2022gophercite} trained \texttt{GopherCite} to produce answers with verbatim quotes from retrieved documents via RLHF, achieving 80\% supported-and-plausible rate on NaturalQuestions, but warned that ``not all claims supported by evidence are true''---even real citations can ground false claims if the source itself is unreliable. These systems introduced the search-then-cite approach that is now standard in commercial LLMs. Production systems, however, may generate URLs from parametric memory rather than actual browsing, and this gap between controlled retrieval and real-world deployment is where citation hallucination emerges.

\paragraph{Attribution and citation evaluation.} \citet{rashkin2023ais} defined Attributable to Identified Sources (AIS), stipulating that NLG output should be verifiable against an independent source document. Their annotation framework includes an ``Article Not Accessible'' category for non-resolving links, but treats it as an edge case rather than quantifying its prevalence. \citet{bohnet2023attributed} formalized attributed QA: even the best retrieve-then-read system achieves only 65.5\% AIS, with a surprisingly low correlation between answer correctness and attribution (Pearson $r = 0.45$). \citet{gao2023rarr} proposed RARR for post-hoc attribution repair---automatically finding web evidence and revising unsupported content---a mitigation approach that assumes real evidence exists for the claim but cannot address fabricated URLs. \citet{gao2023alce} introduced ALCE, the first end-to-end benchmark for LLM-generated inline citations, where even the best models lack complete citation support 50\% of the time on ELI5. These frameworks evaluate whether citations \emph{support claims}, assuming the cited source exists. We test that assumption directly.

\paragraph{Citation generation methods.} \citet{aly_learning_2024} use weakly-supervised fine-tuning (CaLF) with a factual consistency model as a critic, improving citation F1 by 34 points over in-context learning on ALCE-ASQA and outperforming \texttt{GPT-4} by 13 points. \citet{ye_effective_2024} take a different approach with AGREE: they train LLMs to self-ground claims and iteratively retrieve additional evidence for self-identified unsupported statements, gaining 16--37 points in citation recall over prompting baselines. In the long-context setting, \citet{zhang_longcite__2024} show that supervised fine-tuning with sentence-level citation data enables an 8B model to reach citation F1 of 72.0, above \texttt{GPT-4o} (65.6). A consistent finding emerges across these methods: prompting alone does not produce reliable citations, and explicit grounding objectives during training are necessary. All three, however, operate on closed corpora where cited documents are known to exist; they do not address URL fabrication in open-web settings.

\paragraph{Verifiability in generative search.} \citet{liu2023verifiability} audited four generative search engines (Bing Chat, NeevaAI, Perplexity, YouChat) and found that only 51.5\% of generated sentences are fully supported by their citations, despite responses appearing highly informative (average perceived utility 4.50/5). \citet{wu2025sourcecheckup} developed SourceCheckup for medical citation verification, the closest concurrent work to ours. In their evaluation, API-only models (without web access) produce valid URLs only 40--70\% of the time, while RAG-enabled models avoid URL hallucination in their medical evaluation. They also used the Wayback Machine to check whether invalid URLs had ever existed, finding low archived rates, which they interpret as evidence that ``invalid links were indeed hallucinated.'' Our work extends this line of investigation in three directions: First, we span multiple domains rather than medicine alone, revealing domain-dependent variation; Second, we evaluate Deep Research agents, a category SourceCheckup does not cover; and Third, we find that even RAG-enabled systems fabricate 3--13\% of citation URLs, which suggests that the problem is more pervasive than their medical-focused evaluation indicates.

\paragraph{Citation quality beyond support.} The standard binary evaluation (does the source support the claim?) misses important dimensions of citation quality. \citet{xu_citeeval__2025} introduce CiteEval, a principle-driven framework assessing redundancy, credibility, and missing evidence on a Likert scale; their automatic metric achieves Pearson correlation of 0.731 with human judgment, far exceeding NLI-based AutoAIS (0.219). \citet{saxena_generation_time_2025} compare generation-time and post-hoc citation paradigms; human evaluation yields hallucination rates of 41\% and 37\% respectively. and both paradigms produce substantial errors, though retrieval augmentation is the dominant quality driver in either case. \citet{wallat_correctness_2024} disentangle citation \emph{correctness} (whether the cited document supports the statement) from citation \emph{faithfulness} (whether it causally influenced generation): up to 57\% of citations in RAG outputs are ``post-rationalized,'' meaning the model generates from parametric memory and then cites a superficially matching document. Correctness-only evaluation, the standard in the attribution literature, therefore systematically misses unfaithful citations. None of these frameworks account for sources that do not exist at all.

The attribution literature asks ``does the source support the claim?'' Our work asks a more basic question: ``does the source exist?'' A citation can fail by being (a)~nonexistent (our focus), (b)~real but unsupportive (the attribution focus), or (c)~both.

\subsection{Citation hallucination and verification}

\paragraph{Early empirical studies.} \citet{alkaissi2023chatgpt} first documented that \texttt{ChatGPT} fabricates scholarly citations, finding that all five references generated for a medical topic had nonexistent titles, with PubMed IDs that pointed to unrelated papers. When prompted again, \texttt{ChatGPT} returned the same fabricated titles with different years and different mismatched PMIDs. \citet{walters2023fabrication} conducted the first systematic study across 42 multidisciplinary topics: 55\% of \texttt{GPT-3.5} citations and 18\% of \texttt{GPT-4} citations are entirely fabricated. Among non-fabricated citations, 43\% (\texttt{GPT-3.5}) and 24\% (\texttt{GPT-4}) contain substantive errors such as wrong volume, issue, or page numbers. Observe that hyperlinks were \emph{more} likely to appear in fabricated citations (18\%) than real ones (5\%), and that fabricated citations invariably used real journal names with nonexistent article titles.

\paragraph{Domain-specific citation fabrication.} \citet{mugaanyi2024doi} report DOI validity of 70.9\% for natural science citations but only 38.3\% in humanities, with DOI hallucination reaching 89.4\% in the humanities---the closest prior result to our domain-stratified analysis. \citet{dahl2024legal} profiled legal hallucinations across 15,000 federal court cases, with rates from 58\% (\texttt{GPT-4}) to 88\% (\texttt{Llama~2}). and hallucination varies systematically with court hierarchy (Supreme Court lowest, district courts highest), jurisdiction, and case prominence---a frequency-dependent effect consistent with our domain-level findings. In \citet{linardon2025mental}, fabrication rates reach 28--29\% for less-studied mental health topics (binge eating disorder, body dysmorphic disorder) vs.\ 6\% for well-studied ones (major depressive disorder), and that even among real citations, 45\% contain errors such as invalid DOIs. \citet{chelli2024hallucination} report hallucination rates from 28.6\% (\texttt{GPT-4}) to 91.4\% (\texttt{Bard}) for medical systematic review citations; DOI accuracy for non-hallucinated references is only 16--20\%. \citet{aljamaan_reference_2024} develop a Reference Hallucination Score (RHS) for medical chatbots: \texttt{ChatGPT~3.5} and Bing score the maximum (RHS=11/11), while retrieval-augmented tools such as Elicit score the minimum (RHS=1/11). Prompt complexity significantly increases hallucination rates across all systems. The near-zero hallucination in retrieval-augmented tools like Elicit contrasts with our finding that even RAG-enabled systems fabricate 3--13\% of citation URLs. Retrieval reduces fabrication but does not eliminate it at scale.

\paragraph{Large-scale and concurrent work.} \citet{xu2026ghostcite} present the most comprehensive concurrent study (GhostCite): they benchmark 13 LLMs across 40 research domains (375K citations), with hallucination rates from 14\% to 95\% (a 51-percentage-point spread across domains). They also analyze 2.2M citations from 56K published papers: 1.07\% contain invalid citations, with an 80.9\% increase in 2025. Their researcher survey reveals a ``verification gap'': 87\% claim to always verify AI-generated citations, yet 42\% copy-paste BibTeX without checking and 77\% of reviewers do not thoroughly check references. \citet{yuan2026citeaudit} construct a CiteAudit benchmark of 6,475 real and 2,967 fake citations with human-validated labels, and propose a multi-agent verification pipeline achieving 97\% accuracy on generated benchmarks---but only 90\% on real-world cases, where even \texttt{GPT-5.2} drops from 96\% to 33\% F1, The gap between benchmark and real-world detection remains substantial.

\paragraph{Popularity bias and published contamination.} \citet{algaba_large_2024} prompted \texttt{GPT-4}, \texttt{GPT-4o}, and \texttt{Claude~3.5} to suggest references for anonymized in-text citations in 166 ML papers: only 64--68\% of generated references correspond to real papers, and the references that do exist have median citation counts 1,326 higher than ground truth, indicating a strong popularity bias (Matthew effect). \citet{algaba_how_2025} scale this analysis to 274K references generated by \texttt{GPT-4o} for 10K papers across disciplines; existence rates are 40--50\%, and approximately 90\% of existing generated references fall in the top 10\% most-cited papers in their field. Both Algaba studies examine non-retrieval settings where LLMs generate references entirely from parametric memory; our work shows that citation hallucination persists even when models have access to web search, a qualitatively different failure mode. The problem extends beyond model outputs into the published literature: \citet{sakai_hallucitation_2026} audit 17,842 papers from ACL, NAACL, and EMNLP (2024--2025) and identify 295 containing hallucinated references, with a greater than $10\times$ increase from 2024 (0.28\% of papers) to 2025 (2.59\%). The presence of hallucinated references in peer-reviewed proceedings motivates the automated verification pipeline we present in Section~\ref{sec:mitigation}.

\paragraph{Link rot and web content stability.} \citet{zittrain2014perma} found over 70\% of URLs in the Harvard Law Review and 50\% in Supreme Court opinions no longer link to originally cited content. The Pew Research Center found 25\% of webpages from 2013--2023 are no longer accessible \citep{pewresearch2024linkrot}. These baseline link-rot rates contextualize our non-resolving URL findings and motivate our Wayback Machine-based distinction between stale URLs (once real but now non-resolving) and hallucinated URLs (never existed).

\paragraph{Citation hallucination taxonomies and benchmarks.} \citet{ansari_compound_2026} manually coded 100 fabricated citations from 53 NeurIPS 2025 papers, developing a five-category taxonomy: total fabrication (66\%), partial attribute corruption (27\%), identifier hijacking (4\%), placeholder hallucination (2\%), and semantic hallucination (1\%). All 100 exhibited compound failure modes, combining a primary failure with secondary characteristics like semantic plausibility; this layering explains why peer review fails to catch them. Our URL-level verification catches all five categories, since each produces a nonexistent or invalid URL regardless of the specific fabrication mechanism. A different dimension of the problem appears in multi-turn settings: \citet{ram_citation_2025} find that citations mutate substantially between conversation turns across four \texttt{LLaMA} variants, with fabrication rates from 29\% to 86\% even when the topic remains fixed. Our evaluation uses single-turn outputs; the drift finding suggests that citation hallucination may worsen in conversational workflows. On the evaluation side, the CiteME benchmark \citep{press_citeme__2024} evaluates citation attribution (given a text excerpt, identify the cited paper) and finds that frontier LLMs achieve 4--18\% accuracy versus 70\% for human experts; even tool-augmented agents reach only 35\%. This large gap partly explains why models fabricate citations: they lack the ability to reliably ground claims to specific sources. \citet{li_bibagent__2026} construct MisCiteBench, 6,350 expert-validated miscitations across 254 fields, and report near-perfect verification accuracy on accessible sources that drops to 66--80\% when the cited paper is behind a paywall. Our \texttt{urlhealth} tool sidesteps this paywall constraint: checking whether a URL has ever existed requires only HTTP requests and Wayback Machine queries, not access to the cited content.

Most of the above work studies citation fabrication in non-retrieval settings. Our work shows that citation hallucination persists in retrieval-augmented settings (Section~\ref{sec:characterization}), where models fabricate 3--13\% of URLs despite having access to real web search results.

\subsection{Deep research agents}

\paragraph{Architecture and report generation.} Deep research agents build on the reasoning-action paradigm introduced by ReAct \citep{yao2023react}, which showed that interleaving reasoning traces with tool-use actions reduces hallucination from 14\% to 6\% of false positives in QA tasks---a significant reduction, but not elimination. Modern agents can be understood as agentic RAG systems \citep{singh2025agenticrag}, incorporating planning, reflection, and multi-agent collaboration. \citet{shao2024storm} introduced STORM, which generates Wikipedia-like articles with inline citations by simulating multi-perspective expert conversations grounded in internet sources. and the system achieves approximately 85\% citation recall and precision, with remaining errors stemming primarily from improper inferential linking (drawing connections unsupported by sources) and inaccurate paraphrasing rather than source fabrication. STORM is a direct academic precursor to commercial deep research agents; its error analysis shows that citation \emph{faithfulness} failures (content not supported by the source) are distinct from citation \emph{existence} failures (source does not exist), the latter being our focus.

\paragraph{Commercial systems.} OpenAI Deep Research (February 2025)\footnote{\url{https://openai.com/index/deep-research/}} and Google Gemini Deep Research (late 2024)\footnote{\url{https://blog.google/products/gemini/google-gemini-deep-research/}} are among the primary commercial systems whose outputs we evaluate. These agents iteratively formulate search queries, read results, identify knowledge gaps, and synthesize multi-page reports with citations.

\paragraph{Benchmarks.} \citet{du2025drbench} introduced DRBench, 100 PhD-level research tasks with two evaluation frameworks: RACE for report quality and FACT for citation accuracy. The FACT framework evaluates whether the content at a cited URL supports the associated claim, with citation accuracy ranging from 78\% (OpenAI Deep Research) to 94\% (Claude with search). Recall that this evaluation assumes URLs are retrievable---an assumption our work tests directly. \citet{zhong2026draco} evaluate Deep Research agents with DRACO on 100 tasks drawn from production Perplexity usage across 10 domains; citation quality and factual accuracy are the weakest performance axes (the best system achieves only 65\% citation quality and 68\% factual accuracy), and Medicine and Law receive the most penalty-heavy rubrics, consistent with our finding that high-stakes domains are most vulnerable.

Existing deep research benchmarks evaluate report \emph{quality} (accuracy, completeness, coherence) and citation \emph{support} (does the source back the claim?) but do not verify whether cited URLs actually exist. Our work tests a precondition these benchmarks assume: if 3--13\% of URLs are fabricated, citation support metrics systematically overestimate reliability. Our work provides a characterization of citation URL validity across leading commercial LLMs and deep research agents.

\subsection{Citation verification and mitigation tools}

Post-hoc verification of LLM-generated citations has received limited attention as a standalone research direction until recently. The RARR system \citep{gao2023rarr} automatically finds web evidence for LLM claims and revises unsupported content, but it operates at the claim level rather than the citation level, and it assumes that supporting evidence exists somewhere on the web.

\paragraph{Semantic citation verification.} \citet{haan_semanticcite__2025} move beyond existence checking to full-text content verification with SemanticCite, classifying citations into four categories (supported, partially supported, unsupported, uncertain) using fine-tuned lightweight models that achieve 84\% weighted accuracy. \citet{choi_citeguard__2026} frame verification as citation attribution alignment with CiteGuard and achieve 68\% accuracy on the CiteME benchmark, approaching human performance (70\%), and report that LLM-as-judge achieves only 16--17\% recall without tool-augmented retrieval. Both systems require the cited source to be accessible; when the URL itself is fabricated, content-based verification fails silently.

\paragraph{Web archiving and tool ecosystems.} The web archiving community has long studied link rot. \citet{zittrain2014perma} documented reference decay in legal citations and proposed Perma.cc as a solution; \citet{pewresearch2024linkrot} quantified the scale of web content disappearance. Our \texttt{urlhealth} tool operationalizes the Wayback Machine-based verification approach from these studies as a programmatic API. Agent skill ecosystems such as agentskills.io\footnote{\url{https://agentskills.io}} and the Model Context Protocol (MCP)\footnote{\url{https://modelcontextprotocol.io}} provide standardized interfaces for integrating tools into AI agent workflows. and \texttt{urlhealth} is released as an agentskills.io skill, enabling automated citation URL verification within agent pipelines. Our tool operates at the infrastructure level---HTTP requests and Wayback Machine lookups---without requiring additional LLM inference, which distinguishes it from approaches that use LLMs for verification \citep{yuan2026citeaudit,choi_citeguard__2026}.

%% ============================================================
%% APPENDIX C: EXTENDED DOMAIN ANALYSIS
%% ============================================================
\section{Extended domain analysis}
\label{app:domain-analysis}

\begin{table}[ht]
\centering
\small
\begin{tabular}{lrrrr}
\toprule
\textbf{Field} & \textbf{Overall} & \textbf{Claude} & \textbf{Gemini} & \textbf{GPT} \\
\midrule
\multicolumn{5}{l}{\textit{Top 10 by non-resolving URL rate (\%)}} \\
\midrule
Theology & 11.4 & 11.3 & 10.2 & 11.7 \\
Classical Studies & 11.1 & 7.1 & 7.6 & 13.8 \\
Healthcare / Medicine & 10.8 & 17.4 & 5.3 & 7.1 \\
Literature & 10.7 & 9.6 & 7.3 & 11.6 \\
Psychology & 10.2 & 14.7 & 4.6 & 8.6 \\
Biology & 9.6 & 12.8 & 6.1 & 8.1 \\
Visual Arts & 9.4 & 7.7 & 4.1 & 11.9 \\
Linguistics & 8.6 & 6.1 & 4.7 & 10.5 \\
History & 8.3 & 6.8 & 3.9 & 10.1 \\
Military / Law Enforcement & 8.3 & 6.3 & 3.9 & 10.7 \\
\midrule
\multicolumn{5}{l}{\textit{Bottom 5 by non-resolving URL rate (\%)}} \\
\midrule
Economics & 6.2 & 6.1 & 3.9 & 7.4 \\
Engineering \& Technology & 6.1 & 6.0 & 2.5 & 7.2 \\
Journalism & 6.0 & 4.5 & 3.7 & 7.9 \\
Architecture & 5.5 & 5.0 & 2.7 & 6.8 \\
Business & 5.4 & 5.0 & 3.0 & 6.6 \\
\bottomrule
\end{tabular}
\caption{Non-resolving URL rates (\%) by academic field in ExpertQA. ``Claude'' = \texttt{claude-sonnet-4-5}, ``Gemini'' = \texttt{gemini-2.5-pro}, ``GPT'' = \texttt{gpt-5.1}. Top 10 and bottom 5 fields are shown; full results in the supplementary materials.}
\label{tab:expertqa-fields}
\end{table}

\paragraph{Contributing factors.} Several factors may contribute to high rates in Healthcare/Medicine (10.8\% overall) and related life-science fields. Medical web content is frequently updated, moved, or taken down as clinical guidelines change, and healthcare organizations often restructure their web presences. The high publication volume in medical fields may also mean more URLs competing for link stability. Business, Architecture, and Engineering have the lowest non-resolving URL rates, likely because their web content is more stable and less subject to frequent updates.

\begin{figure}[ht]
\centering
\includegraphics[width=\columnwidth]{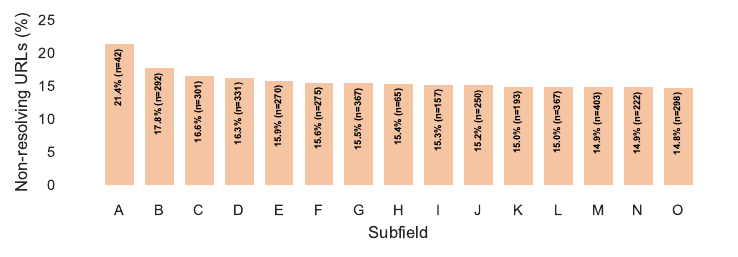}\\[4pt]
\scriptsize
\begin{tabular}{@{}l@{\;\;}l@{\;\;}l@{\;\;}l@{\;\;}l@{}}
\textbf{A:} General practitioner & \textbf{B:} Neuroscience researcher & \textbf{C:} Molecular Microbiology & \textbf{D:} Sales rep in medicines & \textbf{E:} Internal medicine \\[1pt]
\textbf{F:} Reg.\ nurse in trauma & \textbf{G:} Haematology & \textbf{H:} Acute medicine & \textbf{I:} Medical devices & \textbf{J:} Oncology Pharmacist \\[1pt]
\textbf{K:} Laboratory medicine & \textbf{L:} Intensive Care & \textbf{M:} Image guided therapy & \textbf{N:} Neurology & \textbf{O:} Virologist \\
\end{tabular}
\caption{Non-resolving URL rates by subfield within Healthcare/Medicine. The top 15 subfields (sorted by rate) range from 14.8\% (Virologist) to 21.4\% (General practitioner). Subfield labels A--O are keyed in the accompanying table.}
\label{fig:healthcare-subfields}
\end{figure}

\paragraph{Subfield-level analysis.} Figure~\ref{fig:healthcare-subfields} shows that within-field variation can exceed between-field variation. The top 15 Healthcare/Medicine subfields range from 14.8\% (Virologist) to 21.4\% (General practitioner\footnote{Subfield names are derived from self-reported expert descriptions in the ExpertQA dataset~\citep{malaviya2024expertqa} and were not normalized, which accounts for some unconventional labels.}), with Neuroscience researcher (17.8\%, $n=292$) and Molecular Microbiology (16.6\%, $n=301$) among the highest. The spread within this single field (17.1 percentage points from top to bottom) exceeds the gap between many pairs of fields in Table~\ref{tab:expertqa-fields}.

A recurring pattern across high-rate fields is that neuroscience-adjacent subfields rank among the worst within their parent fields: Neuroscience researcher (17.8\%) in Healthcare/Medicine and Neurology (14.9\%) in the same field. In Engineering and Technology, Nuclear Licensing remains an outlier. At the other extreme, fields like Architecture show tight subfield distributions, suggesting that non-resolving URL rates are relatively uniform within stable-content domains.

%% ============================================================
%% APPENDIX D: AGENTIC SELF-CORRECTION DETAILS
%% ============================================================
\section{Agentic self-correction details}
\label{app:self-correction}

\begin{figure}[ht]
\centering
\includegraphics[width=\columnwidth]{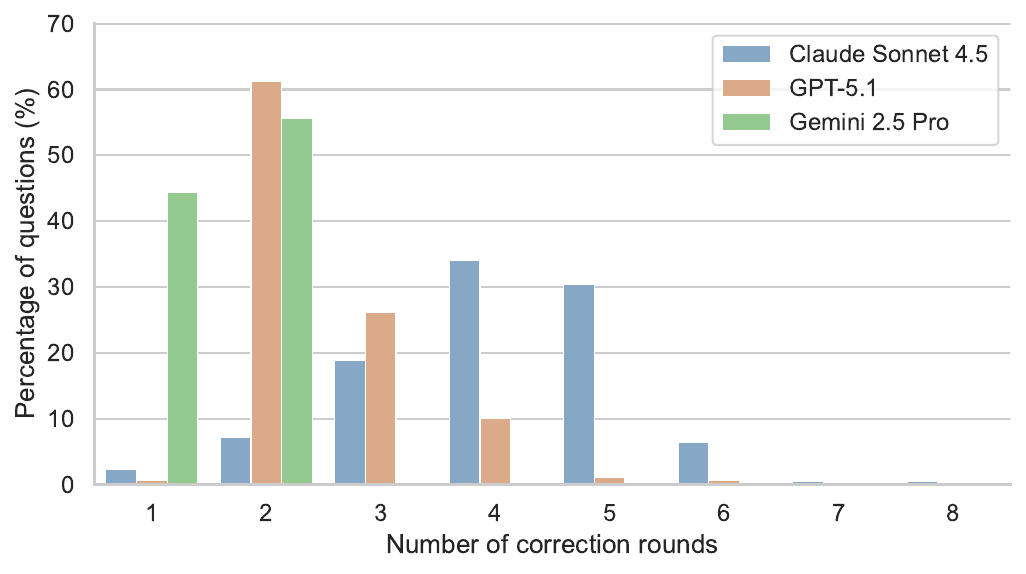}
\caption{Distribution of \texttt{urlhealth} correction rounds per question (435 questions each, 3 models). The three models exhibit distinct self-correction profiles. Gemini~2.5~Pro (green) completes in 1--2~rounds every time: its two-phase architecture (Google Search grounding followed by a single verification turn) caps it at two rounds, and 44\% of questions need only one. GPT-5.1 (orange) clusters at 2~rounds (61\%), with 87\% finishing in 2--3~rounds; the long tail to 6~rounds is rare ($<$2\% of questions). Claude Sonnet~4.5 (blue) concentrates at 3--5~rounds (83\% of questions, median~4), reflecting the cost of verifying its large citation sets (18.4~URLs/question vs.\ 9.7--11.1 for the other models). All three models achieve near-zero hallucinated citations in final output despite these different convergence patterns.}
\label{fig:round-distribution}
\end{figure}

\paragraph{Detailed self-correction behavior.} The three models exhibit distinct behavioral profiles. Gemini completes in 1--2 rounds every time (the two-phase architecture caps it at two rounds), reflecting high initial URL quality that requires little correction. GPT-5.1 clusters at 2~rounds (61.1\% of questions), with 87.3\% finishing in 2--3~rounds and a maximum of 6. Claude clusters tightly at 3--5~rounds (median~4); these rounds are verification-driven, as the model checks its large citation sets rather than replacing hallucinated URLs.

\paragraph{GPT-5-nano failure analysis.} In preliminary experiments with \texttt{gpt-5-nano}, the same pipeline produced a 7.5\% NOT\_LIVE rate in final responses (versus 0.6\% for GPT-5.1), with 48 hallucinated URLs persisting across up to 14 correction rounds. The model called the verification tool but failed to act on its results, repeatedly re-proposing flagged URLs. Replacing \texttt{gpt-5-nano} with \texttt{gpt-5.1}---a model with stronger tool-calling capabilities---reduced the NOT\_LIVE rate to 0.6\%, matching Claude and Gemini.

%% ============================================================
%% APPENDIX E: LIMITATIONS
%% ============================================================
\section{Limitations}
\label{app:limitations}

\paragraph{Wayback Machine coverage.} The Wayback Machine does not archive the entire web. Some hallucinated URLs may be misclassified as stale if they coincidentally match an archived page, and some real URLs may be classified as hallucinated if the Wayback Machine never crawled them. Our hallucination estimates are therefore approximate. However, the net direction of these errors favors under-counting: the headless-browser audit (Appendix~\ref{app:unknown-audit}) finds that 89\% of URLs classified as UNKNOWN by \texttt{urlhealth} are live or blocked-but-accessible, meaning that our pipeline's exclusion of ambiguous URLs removes far more live URLs than dead ones from the non-resolving count.

\paragraph{Temporal sensitivity.} URL liveness checks are point-in-time measurements. Non-resolving URL rates may differ if checked at a different time.

\paragraph{Bot-blocking and rate limiting.} Automated URL checking is subject to false positives from bot-blocking and rate limiting. We mitigate this by excluding HTTP~403 responses and using browser-like User-Agent headers. In the ExpertQA experiment, 403 responses account for 6.6\% (\texttt{gemini-2.5-pro}) to 17.0\% (\texttt{claude-sonnet-4-5}) of all URLs tested, predominantly from major academic publishers: \texttt{sciencedirect.com}, \texttt{researchgate.net}, \texttt{mdpi.com}, \texttt{academic.oup.com}, and \texttt{tandfonline.com} collectively account for over 3{,}100 of the 403 responses. The concentration among publisher domains supports the bot-blocking interpretation, but some fraction of these 403 responses may correspond to genuinely non-resolving URLs, making our non-resolving rates lower bounds. Reddit presents a separate challenge: \texttt{gpt-5.1} generates 15{,}273 Reddit URLs (18.2\% of its ExpertQA citations, compared to $<$1\% for the other models), and Reddit aggressively rate-limits batch requests. Because Reddit URLs that fail batch checks consistently return HTTP~200 when checked individually, we classify all Reddit URLs as alive. This is conservative: some Reddit URLs may be genuinely fabricated. A sensitivity analysis (Appendix~\ref{app:sensitivity}) shows that treating all Reddit URLs as non-resolving raises GPT-5.1's rate from 8.47\% to 26.71\% but does not materially affect the other models.

\paragraph{DRBench model coverage.} We analyze pre-collected data from \citet{du2025drbench}. We did not re-run models and cannot control for prompt variation or API version drift.

\paragraph{ExpertQA model selection.} Only three models were tested on ExpertQA due to API cost constraints. Our results may not generalize to all providers or model sizes.

\paragraph{Non-URL citation hallucinations.} We note fabricated snippets and invented BibTeX entries qualitatively but do not quantify them in this study. Bibliographic metadata verification---validating author names, DOIs, publication venues, and reference existence---is a complementary problem that has received growing attention. FacTool \citep{chern_factool__2023} queries Google Scholar to verify citation existence; CheckIfExist \citep{abbonato_checkifexist__2026} cascades queries across CrossRef, Semantic Scholar, and OpenAlex; CiteAudit \citep{yuan2026citeaudit} proposes a multi-agent verification pipeline; and GhostCite \citep{xu2026ghostcite} provides large-scale evidence of fabricated references. Our study focuses on URL-level citation validity and treats bibliographic metadata verification as out of scope.

\paragraph{Language.} DRBench includes Chinese queries; ExpertQA is English-only. We do not systematically study cross-lingual effects on citation quality.

\end{document}